\documentclass[fleqn,10pt]{wlscirep}
\usepackage[utf8]{inputenc}
\usepackage[T1]{fontenc}
\usepackage{lineno}
\usepackage{amsmath}
\usepackage{subcaption}
\usepackage{bm}
\usepackage{graphicx}
\usepackage{parskip}

\usepackage[format=plain,justification=justified,singlelinecheck=false,font={stretch=1.125,small,sf},labelfont=bf,labelsep=space]{caption}

\def\etal{{\it et al. }}
\newcommand{\ie}[0]{\textit{i.e.}, }

\newcommand{\eg}[0]{\textit{e.g.}, }

\usepackage{chemformula} 
\usepackage{braket}
\newcommand{\DefineAuthor}[2]{%
  \expandafter\newcommand\csname #1note\endcsname[1]{%
    \textbf{\textcolor{#2}{\textbf{#1:} ##1}}}%
  \expandafter\newcommand\csname #1\endcsname[1]{
    \textbf{\textcolor{#2}{##1}}}
  \expandafter\newcommand\csname #1cancel\endcsname[1]{%
    \textbf{\textcolor{#2}{\sout{##1}}}}%
  \expandafter\newcommand\csname #1change\endcsname[2]{%
    \textbf{\textcolor{#2}{\sout{##1} ##2}}}%
  \newenvironment{#1text}{\color{#2}}{\color{black}}
}

\definecolor{dartmouthgreen}{rgb}{0.05, 0.5, 0.06}
\DefineAuthor{LMS}{blue}

\usepackage{xcolor}

\title{Perspective: Towards sustainable exploration of chemical spaces with machine learning}

\author[1,*]{Leonardo Medrano Sandonas}
\author[2]{David Balcells}
\author[3]{Anton Bochkarev}
\author[4]{Jacqueline M. Cole}
\author[5]{Volker L. Deringer}
\author[6,7]{Werner Dobrautz}
\author[8]{Adrian Ehrenhofer}
\author[9]{Thorben Frank}
\author[10]{Pascal Friederich}
\author[11,12]{Rico Friedrich}
\author[13,14]{Janine George}
\author[15]{Luca Ghiringhelli}
\author[16]{Alejandra Hinostroza Caldas}
\author[17]{Veronika Juraskova}
\author[2]{Hannes Kneiding}
\author[3]{Yury Lysogorskiy}
\author[18]{Johannes T. Margraf}
\author[19]{Hanna Türk}
\author[20]{Anatole von Lilienfeld}
\author[21]{Milica Todorovi\'c}
\author[22,*]{Alexandre Tkatchenko}
\author[23,24,*]{Mariana Rossi}
\author[1,25,26]{Gianaurelio Cuniberti}

\affil[1]{Institute for Materials Science and Max Bergmann Center of Biomaterials, TUD Dresden University of Technology, 01062, Dresden, Germany}
\affil[2]{Hylleraas Centre for Quantum Molecular Sciences, Department of Chemistry, University of Oslo, P.O. Box 1033, Blindern, 0315 Oslo, Norway}
\affil[3]{Interdisciplinary Centre for Advanced Materials Simulation (ICAMS), Ruhr-University Bochum, 44780 Bochum, Germany}
\affil[4]{Ray Dolby Centre, Cavendish Laboratory, Department of Physics, University of Cambridge, J. J. Thomson Avenue, Cambridge, CB3 0US. United Kingdom}
\affil[5]{Inorganic Chemistry Laboratory, Department of Chemistry, University of Oxford, Oxford OX1 3QR, United Kingdom}
\affil[6]{Center for Advanced Systems Understanding, Helmholtz-Zentrum Dresden-Rossendorf, Germany}
\affil[7]{Center for Scalable Data Analytics and Artificial Intelligence Dresden/Leipzig, TU Dresden, Germany}
\affil[8]{Institute of Solid Mechanics, TUD Dresden University of Technology, George-Bähr-Str. 3c, 01069 Germany}
\affil[9]{Machine Learning Group, Technische Universität Berlin, 10587 Berlin, Germany}
\affil[10]{Institute of Nanotechnology, Karlsruhe Institute of Technology, Kaiserstr. 12, 76131 Karlsruhe, Germany.}
\affil[11]{Theoretical Chemistry, TUD Dresden University of Technology, Germany}
\affil[12]{Institute of Ion Beam Physics and Materials Research, Helmholtz-Zentrum Dresden-Rossendorf, Germany}
\affil[13]{Federal Institute for Materials Research and Testing (BAM), Unter den Eichen 87, 12205 Berlin, Germany}
\affil[14]{Friedrich-Schiller-University Jena, Institute of Condensed Matter Theory and Optics, Max-Wien-Platz 1, 07743 Jena, Germany}
\affil[15]{Scientific Computing Center, Karlsruhe Institute of Technology, 76131 Karlsruhe, Germany}
\affil[16]{Faculty of Computer Science, TUD Dresden University of Technology, 01062 Dresden, Germany}
\affil[17]{Chemistry Research Laboratory, Department of Chemistry, University of Oxford, Oxford, OX1 3TA, United Kingdom}
\affil[18]{University of Bayreuth, Bavarian Center for Battery Technology (BayBatt), Bayreuth, Germany}
\affil[19]{Ecole Polytechnique Fédérale de Lausanne, Institute of Materials, Lausanne, 1015 Switzerland}
\affil[20]{Department of Chemistry, Chemical Physics Theory Group, University of Toronto, St. George Campus, Toronto, ON M5R 0A3, Canada}
\affil[21]{Department of Mechanical and Materials Engineering, University of Turku, FI-20014 Turku, Finland}
\affil[22]{Department of Physics and Materials Science, University of Luxembourg, L-1511 Luxembourg City, Luxembourg}
\affil[23]{MPI for the Structure and Dynamics of Matter, 22761 Hamburg, Germany}
\affil[24]{Yusuf Hamied Department of Chemistry, Lensfield Road, Cambridge CB2 1EW, United Kingdom}
\affil[25]{Cluster of Excellence CARE, TU Dresden and RWTH Aachen, Germany}
\affil[26]{Cluster of Excellence CeTI, TU Dresden, Germany}

\affil[*]{Corresponding authors:  Leonardo Medrano Sandonas (leonardo.medrano@tu-dresden.de),  Alexandre Tkatchenko (alexandre.tkatchenko@uni.lu), Mariana Rossi (mariana.rossi@mpsd.mpg.de)}

\begin{abstract} 
Artificial intelligence is transforming molecular and materials science, but its growing computational and data demands raise critical sustainability challenges. In this Perspective, we examine resource considerations across the AI-driven discovery pipeline--from quantum-mechanical (QM) data generation and model training to automated, self-driving research workflows--building on discussions from the ``SusML workshop: Towards sustainable exploration of chemical spaces with machine learning'' held in Dresden, Germany. In this context, the availability of large quantum datasets has enabled rigorous benchmarking and rapid methodological progress, while also incurring substantial energy and infrastructure costs.
We highlight emerging strategies to enhance efficiency, including general-purpose machine learning (ML) models, multi-fidelity approaches, model distillation, and active learning. Moreover, incorporating physics-based constraints within hierarchical workflows, where fast ML surrogates are applied broadly and high-accuracy QM methods are used selectively, can further optimize resource use without compromising reliability. Equally important is bridging the gap between idealized computational predictions and real-world conditions by accounting for synthesizability and multi-objective design criteria, which is essential for practical impact.
Finally, we argue that sustainable progress will rely on open data and models, reusable workflows, and domain-specific AI systems that maximize scientific value per unit of computation, enabling efficient and responsible discovery of technological materials and therapeutics.
\end{abstract}
\begin{document}

\flushbottom
\maketitle

\thispagestyle{empty}

\section{Introduction}

Climate change represents one of the most pressing challenges facing humanity and poses a direct threat to the long-term sustainability of modern society. Over recent decades, global energy consumption and carbon emissions have increased dramatically, driven largely by industrialization and the rapid expansion of manufacturing processes. 
In response, regulatory frameworks and international initiatives---including the Paris Agreement\cite{ParisAgreement2015}---are imposing increasingly stringent carbon reduction targets, compelling industries to adopt innovative, energy-efficient solutions that minimize environmental impact without compromising productivity.
Besides \textit{ecological} sustainability, the other two pillars of sustainability are \textit{social} and \textit{economic}. The United Nations Sustainable Development Goals (SDGs) \cite{United2015} provide a comprehensive framework for assessing the societal impact of scientific and technological advances across these dimensions. Molecular and materials science, as key drivers of technological progress, play a central role in this context. When pursued responsibly, the sustainable development of new materials and therapeutics can contribute to multiple SDGs, particularly SDG 7 (Affordable and Clean Energy), SDG 9 (Industry, Innovation, and Infrastructure), SDG 12 (Responsible Consumption and Production), SDG 13 (Climate Action), SDG 11 (Sustainable Cities and Communities), SDG 6 (Clean Water and Sanitation), and SDG 3 (Good Health and Well-being).

Within this broader sustainability landscape, Artificial Intelligence (AI) has emerged as a transformative technology with the potential to reshape industrial processes, optimize energy consumption, reduce waste, and accelerate automation across multiple sectors. In materials science and physical chemistry, AI-driven approaches have already demonstrated remarkable success in accelerating discovery, improving property prediction, and enabling large-scale simulations that were previously computationally prohibitive\cite{huang2023central,keith2021combining,poma25,doi:10.1021/acs.chemrev.4c00572,batatia2025design,Sanchez360,zeniGenerativeModelInorganic2025,zhang2025large,kozinsky_interpretable_bayesian,Vamathevan463,curtarolo2013high,horton_accelerated_2025,BANNIGAN2021113806,doi:10.1021/acs.chemrev.4c00055}.
However, the rapid growth of AI capabilities has also introduced a largely overlooked challenge: the environmental cost of AI itself. The importance of this challenge motivated us to organize a workshop on sustainable AI, which was held in Dresden in the fall of 2025. This perspective aims to highlight the topics we have identified as the most significant for the achievement of sustainable AI in materials science and physical chemistry.

In many areas of application, the concept of ``Green AI" is already discussed \cite{greenai1,greenai2,greenai3}. Its goal is to place sustainability at the core of algorithmic development, hardware utilization, and model deployment. While this paradigm has gained traction in the broader machine learning (ML) community, its adoption within computational materials science and chemistry remains limited, despite the field’s heavy reliance on large-scale simulations and data-intensive workflows. In this regard, methods such as density functional theory (DFT)\cite{dft1,dft2} and \textit{ab initio} molecular dynamics (AIMD)\cite{aimd1,aimd2} have become indispensable for studying technological materials, chemical reactions, and biological responses.
As ML models increasingly augment or replace these traditional simulations, the energy consumption and carbon emission cost of creating training data and subsequently training and deploying models can become more problematic, especially as model complexity and training costs continue to grow.

This trend echoes Jevons paradox~\cite{jevons1865coal}, which states that technological improvements in efficiency can paradoxically lead to increased overall resource consumption. Despite being formulated for the coal economy, it applies to the AI era as well. Recent analyses of applied ML in chemistry and biology reveal a clear upward trend in electricity consumption per publication, driven by longer training times and the use of increasingly powerful GPUs.
In extreme cases, modest improvements in predictive performance incur disproportionate environmental costs. For instance, a reported 28\% reduction in mean absolute error for predicting CO$_2$ storage capacity in metal-organic frameworks was accompanied by a staggering 15,000\% increase in the carbon footprint of model training~\cite{greenmof}.
At a broader scale, the Max Planck Computing \& Data Facility has published statistics on energy consumption and associated CO$_2$ emissions~\cite{mpcdf}, reporting averages (over the past two years) of approximately 37~GWh of electricity use and 14.5~k tons of CO$_2$ emissions per year. 
Similarly, a recent assessment of the Top~500 high-performance computing systems estimates operational emissions of roughly 1.4 million MT CO2e per year \cite{top500}, underscoring the urgent need for more sustainable computational methods.

\begin{figure}[t!]
    \centering
    \includegraphics[width=0.85\linewidth]{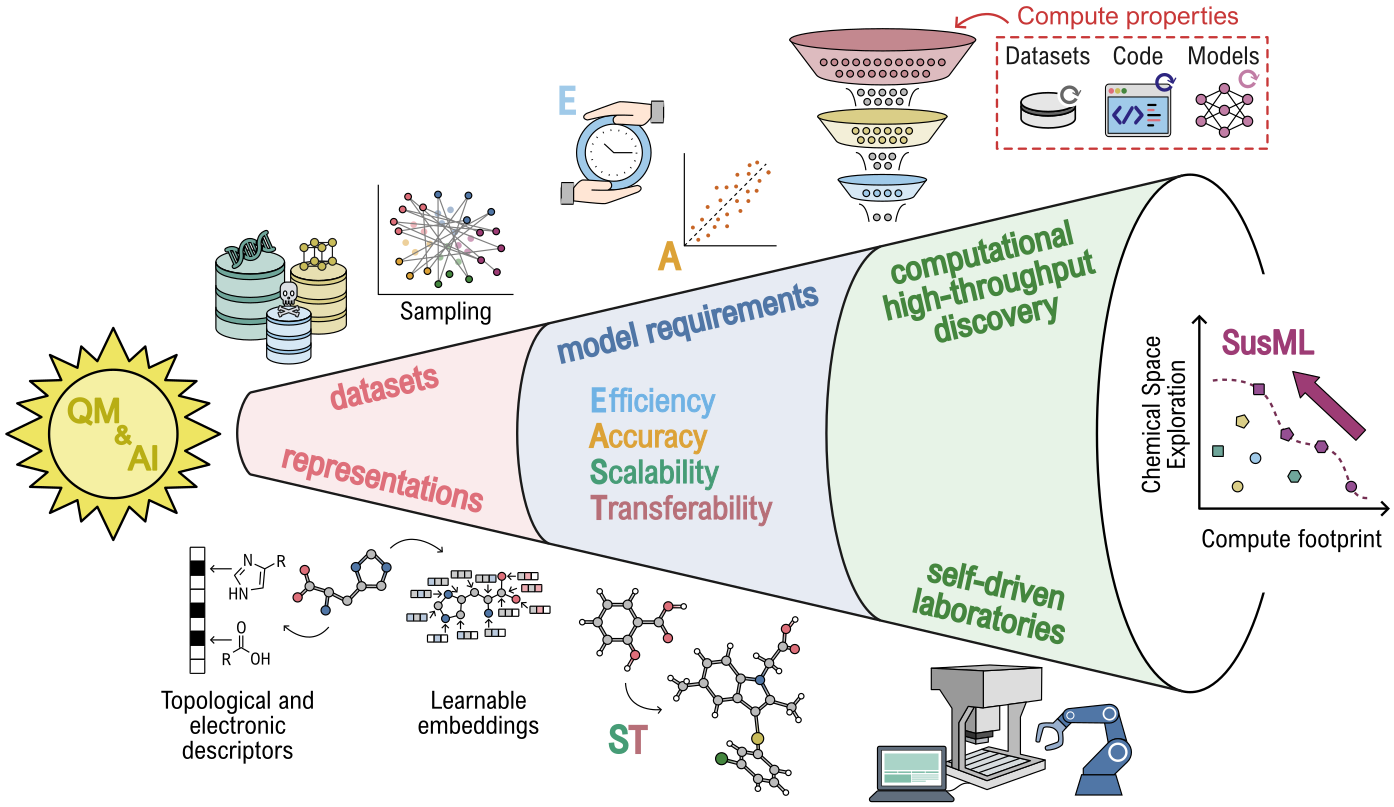}
    \caption{Scheme illustrating the sustainability topics discussed at the SusML workshop (Dresden, Germany) and in this Perspective, spanning the AI-driven discovery pipeline from quantum-mechanical (QM) data generation and model training to automated, self-driving research workflows.}
    \label{fig1}
\end{figure}

Here, we cover the following topics that can lead to more sustainable AI usage (see scheme in Fig. \ref{fig1}), where vast increases of efficiency can be obtained (sufficient to offset model-training costs) and where the community can work towards a sustainable build-up and usage of databases and workflows. 
Since the early works on ML-based property prediction employing simple molecular representations with kernel methods and feedforward neural networks\cite{keith2021combining}, the field has undergone substantial methodological advances.
The development of transformer architectures, graph neural networks, and enriched representations has led to the emergence of general-purpose, often referred to as \textit{foundational}, ML models capable of rapid and accurate property prediction across diverse molecular and materials systems\cite{chen2022universal,deng2023chgnet,kabylda2025molecular,kovacs2025mace,anstine2025aimnet2,batatia_foundation_2025,kim_sevennet_mf_2024, yang2024mattersim,mann2025egret,ple2025foundation,thurlemann2026amp,wood2025family,lysogorskiy2025graph}.
Training on large, openly available datasets has further improved the generalizability and sustainability of these approaches. More recently, quantum-inspired and multimodal representations have been introduced, offering simultaneous gains in accuracy and interpretability (see Fig. \ref{fig2}).

In parallel, generative AI, spanning generative adversarial networks, variational autoencoders, diffusion models, and large language models, has dramatically expanded accessible chemical space, enabling the inverse design of materials and molecules with targeted properties\cite{Sanchez360,zeniGenerativeModelInorganic2025,Kneiding263,turkAssessingDeepGenerative} and accelerating literature-driven discovery workflows\cite{zhang2025large,elsa,Ehrenhofer2025smart_materials_informatics}.
Importantly, recent progress in active learning, uncertainty quantification, and adaptive sampling strategies offers a pathway toward more sustainable AI-assisted discovery and refinement of existing general models. By minimizing the number of required training samples while maintaining accuracy and transferability, these approaches reduce both computational cost and data storage demands. When embedded into high-throughput pipelines, they enable more efficient, automated exploration of chemical space\cite{liu_automated_2025,curtarolo2013high,horton_accelerated_2025,doi.org/10.1002/anie.202200242,BANNIGAN2021113806,doi:10.1021/acs.chemrev.4c00055}.

In the following, we dive deeper into the topics mentioned above, assessing the current state of sustainable ML methodologies in materials science and chemistry, highlighting key shortcomings in existing practices, and discussing emerging strategies for developing better approaches. 

\section{Sustainable property prediction through augmented representations} 

\subsection{Equivariant machine learning force fields}\label{sec:MLIPs}

Great progress has been made in the development of \emph{machine learning force fields} (MLFFs)~\cite{noe2020machine, unke2021machine, keith2021combining}. These models aim to bridge the gap between empirical force fields and quantum-mechanical (QM) methods by learning an approximation to the Born–Oppenheimer potential energy surface (PES) from \emph{ab initio} reference data. Because generating such reference data is computationally expensive, it is highly desirable to construct accurate models using as little training data as possible.
To that end, equivariant MLFFs have emerged as a powerful class of models capable of learning highly accurate interatomic forces from only very few data samples~\cite{schutt2021equivariant, tholke2021equivariant, batatia2022mace, batzner20223, musaelian2023learning, simeon2023tensornet, frank2024euclidean, bochkarev2024graph, batatia2025design}. By encoding physically meaningful geometric priors into their underlying representations, equivariant MLFFs can distinguish subtle interaction patterns that simpler models cannot. This capability enables them to learn more transferable interaction patterns from the same training data. As a result, equivariant MLFFs have been shown to be both more data-efficient and more accurate than their non-equivariant counterparts~\cite{batatia2022mace, batzner20223}. Moreover, they can produce more faithful and accurate dynamics in actual simulations~\cite{fu2022forces, frank2024euclidean}, even with limited training data.
This is further demonstrated by the Graph Atomic Cluster Expansion (GRACE) formalism, which generalizes equivariant message passing neural networks through star- and tree-like graphs while simultaneously providing a formally complete semi-local basis set\cite{bochkarev2024graph}. Thus, employing equivariant representations establishes a connection to classical ML theory, which has shown that richer and higher-dimensional feature spaces are easier to parametrize~\cite{scholkopf1997kernel, vapnik1999nature, braun2008relevant}.

Equivariant MLFFs have also served as an expressive backbone for \emph{pre-trained} MLFFs (also known as universal or foundational or general-purpose models) \cite{chen2022universal, deng2023chgnet, kim_sevennet_mf_2024, yang2024mattersim, mann2025egret, ple2025foundation, anstine2025aimnet2, kovacs2025mace, kabylda2025molecular,batatia_foundation_2025,thurlemann2026amp, wood2025family, lysogorskiy2025graph}. Unlike earlier MLFF versions, which were often trained for one specific system of interest \cite{bartok2010gaussian, chmiela2019sgdml, unke2019physnet, christensen2020fchl, chmiela2023accurate}, these models are trained on large and chemically diverse QM datasets with millions to hundreds of millions of structures (\eg QM7-X \cite{qm7x}, ANI \cite{smith2017ani}, QM9 \cite{Ramakrishnan2014}, MD17/22 \cite{chmiela2017machine,chmiela2023accurate}, DES \cite{donchev2021quantum}, GEMS \cite{gems2024}, QMugs \cite{isert2022qmugs}, Aquamarine \cite{medrano2024dataset}, SPICE \cite{eastman2023spice}, Materials Project\cite{jain2013commentary}, OMat24 \cite{BarrosoLuque2024Open}, Alexandria\cite{cavignac_ai-driven_2025}). Such foundation models can be used directly for a broad set of molecules and materials without any additional training \cite{Pracht2024Efficient,Jakob2025Universally,Gonnheimer2025Beyond}. Alternatively, they can also be fine-tuned on very little domain-specific reference data, usually delivering more accurate results than a model trained from scratch in the low data regime~\cite{batatia_foundation_2025,lysogorskiy2025graph}.

\textbf{Open challenges.} From a sustainability perspective, pre-trained models have the potential to greatly reduce--or even eliminate--the need for system-specific training and data generation. Nonetheless, creating the large datasets and training these models requires a substantial upfront investment in computational resources.
This makes the open availability of these models and datasets essential, since the energy costs involved in their generation can only be amortized through their application across the wider scientific community. A negative side effect of the increasing importance of pre-trained models is that the field can become ``locked-in'' at the levels of theory commonly used for large datasets (\eg PBE for materials), even when other functionals or levels of theory would be more appropriate for a given application. To address this issue, multi-fidelity approaches that combine different levels of theory have been proposed \cite{Cui2025Multi-fidelity}.

Another major sustainability concern for equivariant MLFFs is their computational cost at inference. Their high expressiveness contributes to this cost in several ways. First, they typically have many more trainable parameters than simpler models, such as shallow Behler-Parinello networks or Atomic Cluster Expansion (ACE) models \cite{Behler2007Generalized,Drautz2019Atomic,bochkarev2022efficient}.
More critically, equivariant message passing involves tensor products governed by sparse Clebsch-Gordan coefficients, while modern GPUs are optimized for dense operations. This cost can be mitigated with specialized tensor kernels, bringing equivariant operations close to the efficiency of dense operations \cite{Bharadwaj2025Efficient,10.1145/3731545.3731594}. Alternatively, non-equivariant architectures may offer a route to highly expressive yet computationally efficient MLFFs \cite{10.5555/3666122.3669600}.
Parallelization efficiency is another key factor. Large-scale and long-time simulations can be accelerated by combining efficient MLFFs with parallelization techniques, such as domain decomposition for local MLFFs~\cite{thompson2022lammps} or graph partitioning for message-passing MLIPs~\cite{park2024scalable,han2025distmlip}, substantially reducing computational requirements. Finally, large general-purpose models can be distilled into smaller and more efficient versions without significant loss of accuracy \cite{Gardner2025Distillation}.

\subsection{Property prediction with multi-modal representations}

\subsubsection{Quantum-informed representations for drug discovery}

Computational-aided drug discovery (CADD)\cite{cadd,BANNIGAN2021113806} critically depends on expressive molecular representations to build predictive models that generalize beyond the training data. Among the most relevant endpoints in this context are absorption, distribution, metabolism, excretion, and toxicity (ADMET) properties, which strongly influence drug safety and clinical success. Suboptimal ADMET profiles remain a major cause of drug attrition \cite{cook2014astrazeneca,dossantos2022drug}, and late-stage failures are particularly costly \cite{kola2004can}. Consequently, the early identification of unfavorable ADMET endpoints has become a central objective in CADD.

This challenge has stimulated extensive research within Quantitative Structure–Activity/Property Relationship (QSAR/QSPR) frameworks, where machine learning (ML) methods are now widely adopted. Deep learning approaches can directly process molecular string representations such as the Simplified Molecular Input Line Entry System (SMILES) \cite{weininger1988smiles}, treating them as sequences through masked language modeling \cite{ahmad2022chemberta} or transformer-based architectures \cite{aksamit2024hybrid,born2023chemical}. Alternatively, models may operate on molecular graphs derived from the Lewis structures encoded in these strings.
This graph-based paradigm has been extensively explored in architectures such as Attentive FP \cite{xiong2019attentivefp} and Chemprop \cite{heid2023chemprop}, among others \cite{tang2020self,ghanavati2024machine,zalte2025rigr,decarlo2024smile}. These approaches combine atomic and bond features within two-dimensional molecular graphs to predict physicochemical and biological properties, including lipophilicity, solubility, and toxicity. Although such models often achieve strong predictive performance and incorporate mechanisms for interpretability (\eg attention weights), their explanations remain limited. Moreover, they tend to struggle in challenging regimes characterized by data scarcity, activity cliffs, and non-smooth structure-property relationships---conditions frequently encountered in ADMET prediction tasks \cite{xia2023limitations,jiang2021graph}.

In this context, traditional ML methods, such as tree-based models and support vector machines, often achieve robust performance when paired with chemically and physically meaningful molecular descriptors. Unlike purely topological representations, these descriptors encode information beyond molecular connectivity. Examples include substructure-based fingerprints \cite{pattanaik2020fingerprints}, geometric representations \cite{keith2021combining,huang2021ab}, and selected physicochemical features such as molecular weight \cite{chtita2021qsar} or Fukui functions \cite{stuyver2022quantum}.
The predictive relevance of these descriptors suggests that further progress in ADMET modeling may require a shift toward systematically enriching molecular representations. One principled approach is to incorporate QM features. Quantum-inspired descriptors explicitly encode electronic structure information, such as molecular orbital energies and charge distributions, that govern intermolecular interactions and ultimately determine macroscopic physicochemical behavior and biological activity.
Such descriptors can be computed using accurate density-functional theory (DFT) frameworks \cite{li2024quantum,tortorella2021combining,shimakawa2024qmex} or more computationally efficient semi-empirical approaches \cite{abarbanel2024qupkake,macorano2025quantum}. Beyond serving as additional input features, QM descriptors can support pre-training strategies in deep learning models \cite{fallani2024quantum}, enhance the predictive power of existing topological and geometric representations, and improve interpretability by grounding predictions in physically meaningful quantities. This is particularly relevant for ADMET endpoints such as permeability and solubility \cite{bose2026quantum,zhang2025quantum}.

\begin{figure}[t!]
    \centering
    \includegraphics[width=0.85\linewidth]{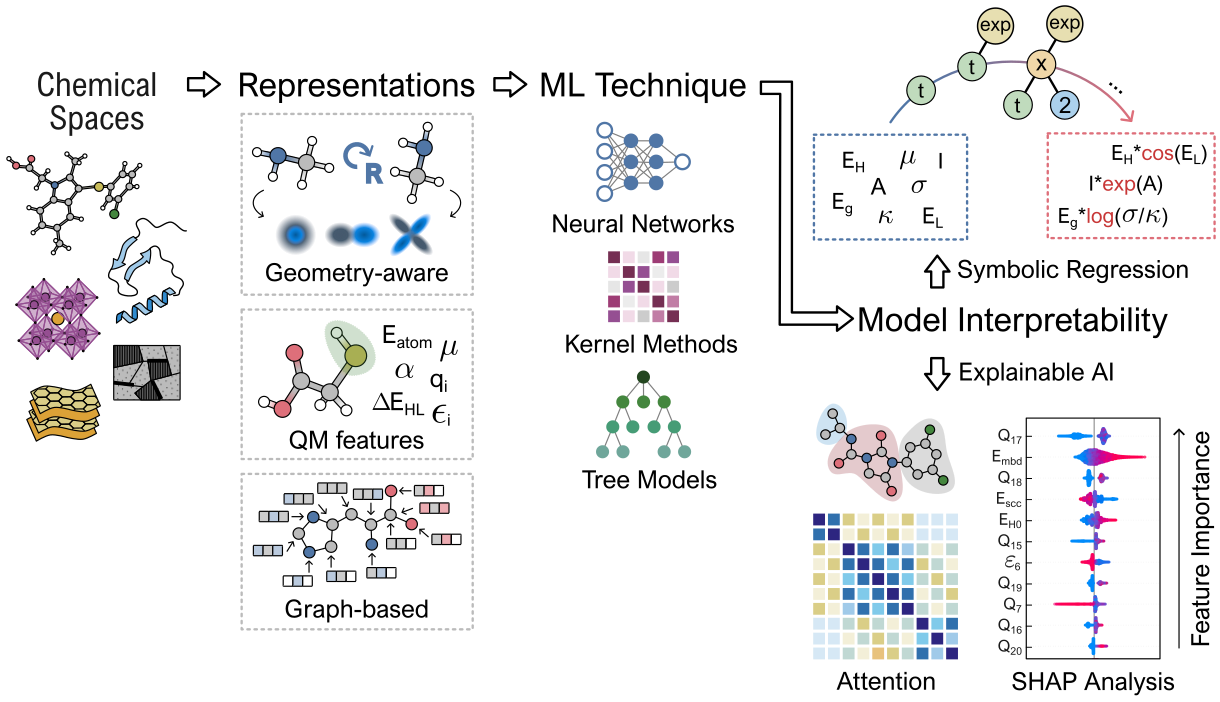}
    \caption{Standard workflow for developing predictive models of molecular and materials properties, combining quantum-inspired representations with machine-learning (ML) techniques (\eg neural networks, kernel methods, and tree-based models). Certain ML models can be made interpretable through additional explainability techniques, such as symbolic regression, attention mechanisms, and SHAP (SHapley Additive exPlanations) analysis..}
    \label{fig2}
\end{figure}

A recent example is the QUantum Electronic Descriptor (QUED) framework \cite{hinostroza2026assessing}, which combines electronic features computed at the semi-empirical DFTB3+MBD level\cite{dftb+} (\eg atomic charges, orbital energies, and energy components) with low-cost geometric descriptors such as Bag of Bonds, SLATM, and SOAP\cite{keith2021combining}. QUED demonstrates that integrating electronic and geometric information can yield accurate predictions of toxicity and lipophilicity while enabling explainable AI analyses that identify the molecular properties most relevant to a given regression task (see ``Explainable AI'' block in Fig. \ref{fig2}).
Moreover, QUED can be further extended to consider learnable embeddings derived from graph neural networks or equivariant ML interatomic potentials, such as MACE \cite{kovacs2025mace} and SO3LR \cite{kabylda2025molecular}. These models, which encode symmetry-aware representations of molecular structure, have already shown promising performance in ADMET prediction tasks \cite{cremer2023equivariant}, suggesting a pathway toward hybrid representations that combine physical grounding with data-driven flexibility.

\textbf{Open challenges.} Therefore, the systematic generation of QM datasets tailored to ADMET prediction represents a crucial step toward more sustainable CADD. By introducing physically grounded information that enhances model robustness and generalization--particularly in low-data regimes--QM descriptors can improve data efficiency while remaining reusable across multiple endpoints and modeling tasks.
A promising direction is this regard is the integration of ML-augmented electronic structure methods\cite{wei21,Guoqing22,huang2023central,equidtb,li_critical_2025}, which can substantially reduce the computational cost of generating such descriptors and enable scalable workflows. In this way, QM-based features can support more sustainable and trustworthy ADMET prediction pipelines.

Importantly, electronic properties are inherently dependent on the molecular structure used in their calculation. Consequently, QM-based representations should be coupled with rigorous conformational sampling, enabling the characterization not only of a single optimized geometry but of an ensemble of thermodynamically relevant conformers. In this context, pre-trained generative AI models may further facilitate conformational exploration, potentially increasing the diversity and coverage of biologically relevant chemical space \cite{Fallani24,xu25,Volokhova24}---a central bottleneck for many ML methodologies. Accounting for conformational variability in the prediction of ADMET endpoints remains largely unexplored, yet it could exert an impact comparable to that observed in the development of ML interatomic potentials and in the generative design of drug-like molecules.

\subsubsection{Parsimonious models for materials informatics}

As in CADD, a central challenge in materials informatics is predicting complex collective properties, such as thermal conductivity, optical band gap, elastic moduli, or thermodynamic stability, from relatively simple descriptors derived from atomic-scale information, including elemental attributes such as electronegativity, ionic radii, and valence orbital energies \cite{butler2018machine,ramprasad2017machine,ghiringhelli2015bigdata}. This paradigm, underlying the Materials Genome Initiative and related data-driven discovery efforts, seeks to compress high-dimensional physical information into compact predictive models that are both interpretable and practically useful for materials design \cite{jain2013commentary}.
A fundamental challenge in this context is the vastness of materials space\cite{curtarolo2013high}. Existing computational and experimental databases sample only a minute fraction of this space, typically comprising at most thousands to tens of thousands of entries, which is negligible relative to the underlying chemical and structural dimensionality \cite{schmidt2019recent}. As a consequence, supervised learning in materials informatics operate in a sparse-data regime, where conventional ML approaches that rely on dense sampling are prone to overfitting and poor extrapolation \cite{bishop2006pattern}.
This intrinsic sparsity motivates the development of learning methods that remain reliable under limited data. Sparse regression approaches have therefore attracted considerable attention in materials informatics because they balance predictive accuracy with interpretability \cite{tibshirani1996regression}.By augmenting least-squares regression with penalties on the number of active features, these methods favor models that depend on only a small subset of candidate descriptors. However, they are fundamentally restricted to linear combinations of predefined features. Nonlinear expressiveness can be increased by introducing interaction terms or polynomial expansions, but the resulting feature spaces grow combinatorially, quickly becoming computationally intractable and reducing interpretability \cite{hastie2009elements}.

Symbolic regression and classification methods address this limitation by constructing models as explicit analytical expressions obtained by combining primary features with a predefined set of mathematical operators (see ``Symbolic Regression'' block in Fig. \ref{fig2})\cite{koza1992genetic}. In this framework, the functional form of the model is not fixed a priori but is discovered as part of the learning process. 
Historically, symbolic regression has relied on stochastic global optimization strategies, most notably genetic programming \cite{koza1992genetic}. While such approaches are expressive and flexible, they often suffer from limited scalability, sensitivity to hyperparameters, and a lack of reproducibility due to their stochastic nature \cite{schmidt2009distilling}. 
More recently, symbolic regression has been combined with deep learning techniques, including graph-based neural representations, to guide the search over expression space and improve efficiency \cite{cranmer2020discovering}.

To combine the expressiveness of symbolic regression with the determinism and data efficiency of sparse regression, the Sure Independence Screening and Sparsifying Operator (SISSO) framework was introduced \cite{ouyang2018sisso}. SISSO reformulates symbolic regression as a compressed sensing problem, enabling systematic exploration of extremely large descriptor spaces under strict sparsity constraints \cite{donoho2006compressed}. Subsequent methodological advances and software developments have further improved its scalability and robustness \cite{purcell2023sissopp}.
Models learned by SISSO are parsimonious by construction, facilitating efficient and reliable prediction of materials properties. Because the number of active descriptors is explicitly limited, accurate models can often be obtained from datasets containing only tens to hundreds of samples, contrasting sharply with universal approximator approaches such as deep neural networks \cite{goodfellow2016deep}. Moreover, once a SISSO model is learned, its evaluation cost is negligible, as it involves only the direct computation of a small analytical expression from known primary features \cite{ghiringhelli2017learning}. These characteristics make SISSO particularly attractive for high-throughput screening and applications where interpretability and physical insight are essential.
The practical utility of SISSO has been demonstrated across diverse materials problems. For example, Bartel \textit{et al.}, who used SISSO to derive a revised tolerance factor for predicting the stability of perovskite structures \cite{bartel2019new}. 
Other studies have produced compact analytical models for temperature-dependent Gibbs free energies \cite{bartel2018physical}, hardness as a function of elastic moduli \cite{TANTARDINI2024102402}, thermodynamic properties of intermetallics, structural classification in perovskites, and pseudocapacitance formulas for MXenes  \cite{WANG2023232834}.

\textbf{Open challenges.} Several key directions remain for the development of sparse symbolic learning in materials informatics. A major challenge is the systematic incorporation of descriptors capturing local atomic order. Many current symbolic models rely primarily on composition-based or averaged features, whereas numerous properties depend sensitively on short- and medium-range structural motifs \cite{bartok2013representing}. Incorporating such information requires local structural descriptors based on symmetry functions, smooth overlap of atomic positions, or graph representations \cite{himanen2020dscribe}. Within a sparse symbolic framework, this must be achieved without uncontrolled feature growth to preserve tractability, interpretability, and computational efficiency---an important consideration for sustainable large-scale screening \cite{schmidt2019recent}.
A closely related frontier is the explicit integration of physics priors, aligning sparse symbolic learning with physics-informed ML \cite{raissi2019physics}. Embedding constraints such as conservation laws, symmetries, dimensional consistency, and asymptotic behavior directly into the symbolic search space can reduce model complexity, improve generalization, and lower data and computational requirements \cite{cranmer2020discovering}. This approach is particularly promising for sustainable materials discovery, where reliable predictions must often be made from limited experimental data and where physically grounded models can help identify environmentally benign compositions and energy-efficient functionalities.

In summary, sparse symbolic learning (\textit{e.g.} SISSO) offers a powerful strategy for modeling in the sparse-data regime. By combining large yet controlled descriptor spaces with strict sparsity enforcement and deterministic optimization, it yields compact, interpretable models that capture essential structure–property relationships while minimizing data and computational demands. These characteristics make the approach attractive for sustainable materials research, enabling efficient exploration of vast design spaces with reduced experimental and computational overhead.

\begin{figure}[t!]
    \centering
    \includegraphics[width=1\linewidth]{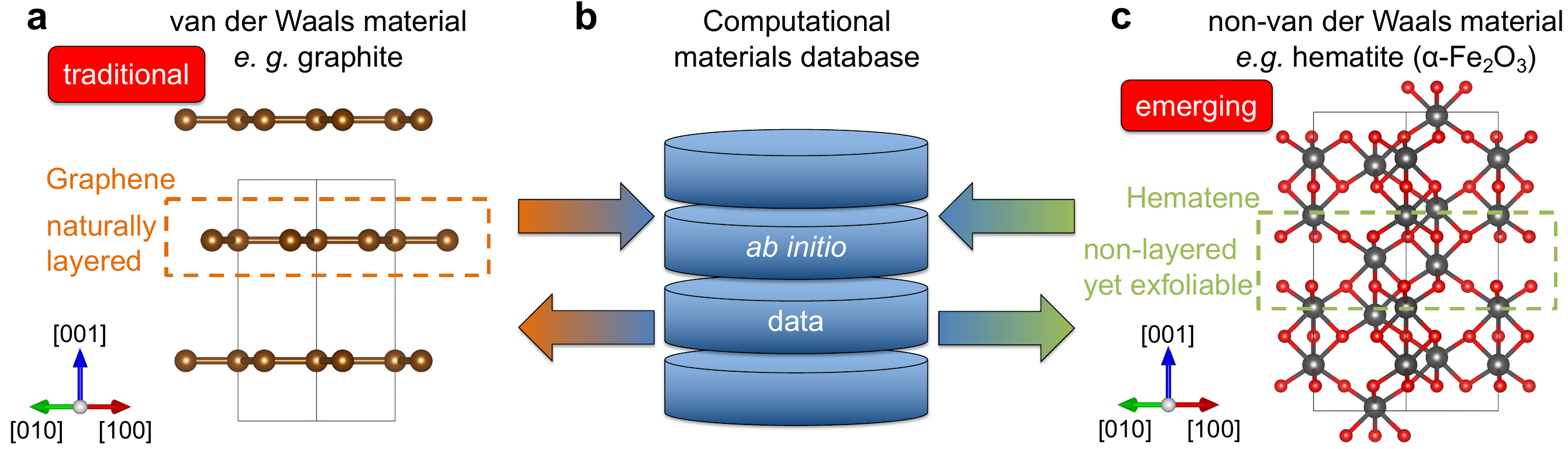}
    \caption{\textbf{Data-driven discovery of traditional \emph{vs}. emerging 2D materials.}
	(\textbf{a}) 2D materials such as graphene are traditionally exfoliated from naturally layered vdW compounds such as graphite.
	Large computational materials databases (\textbf{b}) have already fueled the discovery of several thousand such systems.
	For the emerging class of non-vdW 2D materials derived from non-layered crystals such as hematite ($\alpha$-Fe$_2$O$_3$) (\textbf{c}), data-driven research and ML based modelling is still at an early stage.
	Colors: C brown, Fe gray, and O red.}
    \label{fig:Rico_Friedrich}
\end{figure}

\subsection{Accurate electronic structure for materials discovery}

\subsubsection{Data-driven design of non-van der Waals two-dimensional materials}

Two-dimensional (2D) materials are an important class of nanoscale compounds with a large chemical space, where, due to quantum confinement of electrons within a plane, enhanced or novel properties and potential functionalities can emerge.
As such, these nano-systems can exhibit metallic, semiconducting, insulating, as well as magnetic and topological character of interest for \emph{e.g.} electronic and catalytic applications. 
Traditional 2D compounds are derived from layered crystals held together by weak van der Waals (vdW) forces for which single graphene sheets obtained from bulk graphite are the prime example~\cite{Novoselov_Science_2004} (see Fig.~\ref{fig:Rico_Friedrich}a).
Data-driven and, in particular, ML approaches have been used extensively to widen the chemical space of these 2D systems and to discover and design new candidates applying, for instance, deep generative models \cite{Mounet_AiiDA2D_NNano_2018,Song_ACS_ApplMatInt_2021,Lyngby_npjcm_2022,Wang_2D_Mat_2023,Lyngby_2DMats_2024,Wang_JPCL_2024}.
This was largely enabled by big materials databases dedicated to 2D systems \cite{Mounet_AiiDA2D_NNano_2018,Sorkun_NPJCM_2020,Zhou_SciDat_2019,Haastrup_C2DM_2DM_2018,Gjerding_2DMats_2021,Lyngby_npjcm_2022,Song_ACS_ApplMatInt_2021} with at least several thousands of entries (Fig.~\ref{fig:Rico_Friedrich}b) that had already been built up over several years before based on high-throughput \emph{ab initio} calculations providing the needed data stock for parametrizing the models.

More recently, however, the successful experimental exfoliation of so-called non-vdW 2D sheets from ionicly or covalently bonded non-layered crystals came as a surprise -- shattering our intuitive understanding of exfoliation of crystals.
The first representatives were mainly derived from oxidic mineral ores such as WO$_3$, $\alpha$-Fe$_2$O$_3$ (hematite), and FeTiO$_3$ (ilmenite), leading to the associated 2D sheets such as hematene and ilmenene~\cite{Guan_AdvMat_2017,Puthirath_Balan_NNANO_2018,Puthirath_Balan_CoM_2018} (see Fig.~\ref{fig:Rico_Friedrich}c).
These were followed over the past years by a few dozen other non-vdW 2D systems generated by a variety of experimental techniques~\cite{Balan_MatTod_2022,Kaur_AdvMat_2022,Jiang_NatSyn_2023}.

This emerging new class of 2D nanoscale compounds is particularly interesting because of its qualitatively new properties -- complementing the established space of layered vdW 2D systems.
While traditional 2D materials are often chemically inert with anions at their surfaces, such as in the case of MoS$_2$ sheets, non-vdW systems are intrinsically (re-)active due to their cation surface termination, frequently leading to dangling bonds and surface states upon exfoliation~\cite{Puthirath_Balan_NNANO_2018,Friedrich_NanoLett_2022,Barnowsky_AdvElMats_2023}.
This can give rise to favorable catalytic properties, as demonstrated for potential water splitting applications~\cite{Puthirath_Balan_CoM_2018}, and also to distinct magnetic states, that can even be controlled by surface chemistry approaches such as passivation~\cite{Barnowsky_NanoLett_2024}.
Stacking these sheets into non-vdW heterostructures leads to strong interfacial bonding, giving rise to strong electronic and magnetic coupling, hybrid interface bands as well as pronounced moir{\'e} property variations upon twisting\cite{Nihei_ARXIV_2025}.

From the computational side, previously developed descriptors that enable data-driven investigations tailored to vdW-bonded layered crystals are obviously not applicable to identify novel non-vdW 2D candidates.
First attempts in extending the materials space were made by identifying an intrinsic structural weakness in the hematite and ilmenite crystal structures, promoting exfoliation, such that all materials with the same structural prototype can be expected to exhibit similar cleavage properties.
This approach led to the identification of about three dozen new non-vdW 2D candidates\cite{Friedrich_NanoLett_2022,Barnowsky_AdvElMats_2023}.
Other approaches based on explicit electronic structure calculations~\cite{Jia_npjcm_2021,Hu_NatComm_2023,Ono_PRB_2025,Zhong_npjcm_2025} are not restricted to particular structural features but are prohibitively expensive to execute for all possible exfoliation planes of all possible or known (inorganic) materials.
The usage of similar generative models as in the case of traditional 2D materials is obviously hindered by the scarcity of data for non-vdW 2D systems and a lack of general understanding of what promotes the exfoliability of these compounds.

\textbf{Open challenges.} This scenario thus calls for invoking sustainability for developing proper ML models from the perspective of data efficiency.
At the same time, the models need to realize the meaningful identification of the preferentially exfoliable 2D sheets. 
A viable step in this direction is to deliberately include specific functional forms and physical constraints during the large-scale fitting procedure.
To outline possible non-vdW 2D materials from all possible bulk compounds, this approach was recently employed by deriving exfoliable planes and 2D subunits from a dedicated potential parametrization focused on bonding strength between the atoms~\cite{Barnowsky_ARXIV_2025}.
The developed exfoliation and cleavage (XCP) potential was constructed and fitted by explicitly making use of analytical building blocks in the potential parametrization 
-- thereby strongly reducing the number of parameters and hence the required data for accurate fitting compared to established ML interatomic potentials.
The extractable sheets can be identified by iteratively deleting weak bonds or by evaluating the bonding energy per unit area.
The approach proved to be highly successful by outlining 37,208 cleavable surfaces and potential non-vdW 2D sheets from all materials within the AFLOW-ICSD~\cite{Esters_CMS_2023} repository. 
It also correctly identified almost all experimentally realized systems.
This scheme might, to a certain extent, be seen as a template as the development of data-efficient and physically interpretable approaches is an ongoing challenge.
In several fields other than just non-vdW 2D systems, the lack of data in conjunction with complex scientific questions calls for proper, tailored, sustainable ML approaches.

\subsubsection{High-accuracy electronic structure \& correlated materials}
	
While ML models have shown strong performance for ground-state properties of well-behaved systems\cite{Reiser2022, Nandy9927, Shao2023, Ramakrishnan2015}, many chemically and technologically relevant compounds, including open-shell molecules, transition-metal complexes, excited states and strongly correlated materials, present challenges that go beyond standard single-reference descriptions\cite{Unke2021, deng2023chgnet, Simeon2025}. In these systems, ground and low-lying electronic states with different spin, charge, or correlation strength may be energetically competitive and difficult to distinguish using standard approximations\cite{Morosan2012, LiManni2020}. Such electronic ambiguity is not apparent from structure alone, yet directly influences observables such as total energy, spectra, and chemical reactivity.

ML models can achieve remarkable accuracy for a range of molecular and materials properties\cite{Allen2022, Zubatyuk2019}, but modeling strongly correlated systems remains challenging. These regimes involve, e.g., near-degeneracies or multiconfigurational character. Reliable prediction in such cases requires high-fidelity reference data and physically meaningful features that reflect the inherent electronic correlation. These features can include total spin and charge (when not externally fixed), as well as more subtle indicators of multiconfigurational character such as fractional orbital occupations or entanglement metrics\cite{Unke2021, King2025, Gee2025, deng2023chgnet}. In addition to explicit descriptors, signs of electronic complexity can also be indicated by solver behavior: sensitivity to basis set, instability with respect to active space, or large discrepancies between lower- and higher-level methods may all indicate strong correlation or multireference character.
Such information is difficult to obtain from low-cost methods alone and typically requires high-accuracy solvers with controlled treatment of correlation. 
Embedding these physically grounded quantities into ML pipelines can extend predictive reliability in regions where approximate methods are likely to break down.

High-accuracy methods such as density matrix renormalization group\cite{White1992, Schollwck2011}, auxiliary-field quantum Monte Carlo\cite{Blankenbecler1981, Ceperley1977, Motta2018}, full configuration interaction quantum Monte Carlo\cite{Booth2009, Guther2020, Dobrautz2019}, or neural quantum states\cite{Carleo2017} and other high-accuracy ML approaches\cite{Pfau2020, Hermann2020} provide controlled access to strongly correlated electronic regimes. 
Although computationally demanding, they offer benchmark-quality reference data in electronically complex systems, enable the extraction of physically meaningful quantities that may inform multi-modal machine learning models, and serve as solvers of last resort when standard approximations fail.

Quantum computing on the other hand holds significant promise for advancing electronic structure theory, offering a path to scalable and systematically improvable treatments of strongly correlated quantum systems\cite{Schleich2025,  Alexeev2024, Dobrautz2024}. Novel approaches such as (variational) quantum imaginary time evolution\cite{McArdle2019, Motta2019, Fitzek2024, Magnusson2024}, sample-based quantum diagonalization\cite{Sugisaki2025, RobledoMoreno2025}, and quantum Krylov methods\cite{Yoshioka2025} operate directly in the Hilbert space of the many-electron wavefunction and are naturally suited to capturing quantum correlation. While currently limited by hardware constraints, they are among the most promising strategies for extending high-accuracy solvers to system sizes beyond the reach of classical methods.

\textbf{Open challenges.} These high-accuracy methods clarify where representations must encode the relevant physics explicitly, and where learned approximations cannot compensate for missing electronic information. However, from the standpoint of computational sustainability, the goal is not to replace high-level quantum chemical methods for strong correlation entirely with ML, but to deploy the computationally costly methods only selectively where they are truly needed. Multi-modal representations that reflect key aspects of the electronic structure, such as frontier orbital degeneracy, spin contamination, or strong orbital entanglement, can help identify cases where low-cost approximations are likely to fail.
Such information enables hierarchical workflows, where fast methods and ML surrogates are applied broadly, while high-accuracy solvers are reserved for electronically complex cases. In this setting, physical indicators of electronic complexity support smarter allocation of computational resources.

Many widely used benchmarks and datasets focus on chemically simple, single-reference systems\cite{Ramakrishnan2014, Ramakrishnan2015, qm7x}. Electronically complex regimes, indicated by correlation and/or near-degeneracies, are less represented. Extending ML into these domains will require deliberate efforts to build datasets and representations that reflect their specific challenges. Methods developed in high-performance and quantum computing, including variational wavefunction approaches such as neural quantum states, may support this effort. Their value lies not in generating large-scale datasets, but in providing interpretable features or high-fidelity reference points for electronically complex cases.
The central question is what information a model must encode to predict the correct physics. Framing the problem this way redirects attention toward physically grounded inputs that enable accurate predictions in strongly correlated systems, rather than relying on model size or architecture alone.

\section{Expanding chemical space with sustainable generative and extractive AI models} 

\subsection{Generative frameworks for inverse design}

\begin{figure}[t!]
    \centering
    \includegraphics[width=1\linewidth]{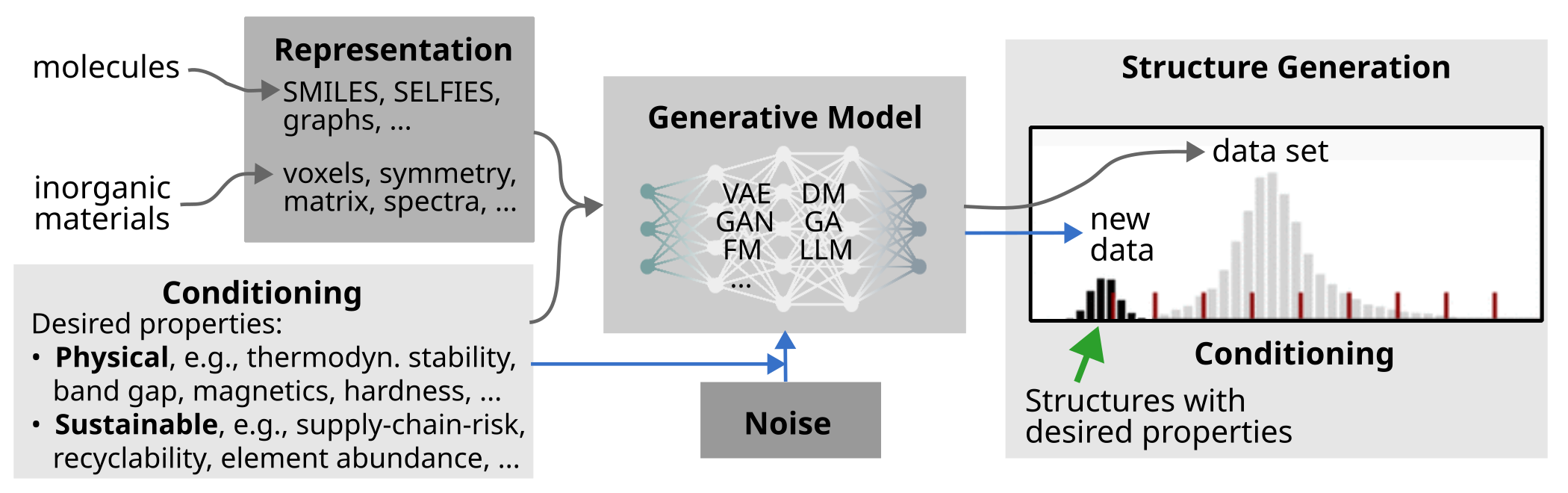}
    \caption{\textbf{General scheme of structure generation with generative models.} Generative models are trained to learn the underlying distribution of a dataset of molecules or materials encoded in a representation, together with their conditioning label on desired properties (gray arrows). From a trained model, new data with the desired properties can be sampled from noise, along with the respective conditioning (blue arrows). The included histogram is reproduced from Ref. [\citenum{turkAssessingDeepGenerative}].}
    \label{fig:GMs}
\end{figure}

Generative AI is establishing itself as a powerful tool in the chemical and materials sciences, enabling the inverse design of molecules\cite{Sanchez360, Anstine8736} and materials\cite{Cheng2026, metni_generative_2025, wangLeveragingGenerativeModels2025} with models that combine efficient ML algorithms with large and complex data (see Fig. \ref{fig:GMs}). By conditioned generation on specific target properties, these models allow for breaking the trial-and-error cycle of research, accelerating materials discovery through  fast and targeted exploration of the chemical space. They promise to do this more efficiently than well-established methods at lower computational cost by directly proposing materials with target properties\cite{renInvertibleCrystallographicRepresentation2022a}, yielding higher compositional and structural diversity than high-throughput screening, being more targeted and flexible than random structure search, and having better structure generation efficiency than crystal structure prediction\cite{yanStructurePredictionMaterials2023, zeniGenerativeModelInorganic2025}.

\subsubsection{Generative AI for molecules}

For molecules, this perspective focuses on transition metal complexes (TMCs), in which one or few metal centers are supported by a set of ligands. Compared to the chemistry of small organic molecules, in which AI-driven drug discovery has been particularly successful\cite{Vamathevan463}, data for TMCs is less abundant and can exhibit significant biases. For example, the Cambridge Structural Database\cite{Groom171} is a highly valuable resource but only contains data for experimentally synthesized TMCs. These difficulties have oriented the focus of ML for TMCs towards data-efficient methodologies, which are inherently more sustainable. AI-valuable data for TMCs can be computed with HPC resources, but the significant size and complex electronic structure of these compounds demand treatment at high, expensive levels of theory. In this framework, high-throughput virtual screening is not a sustainable alternative to AI, as it quickly leads to prohibitively high computational costs.

From the perspective of sustainability, the interest in generative AI for TMCs is also motivated by the intrinsic value of these compounds in the field of green chemistry\cite{Moran9089}. 
TMCs are central to catalysis, enabling more sustainable chemical processes by reducing energy consumption and waste, while underpinning key industrial reactions such as hydrogenation\cite{filonenko2018catalytic} and olefin polymerization\cite{deng2022late}, as well as emerging strategies like CO$_2$ hydrogenation for greenhouse gas capture and utilization\cite{das2022recent}.
In line with this, TMCs have also found successful application as chromophores in light-harvesting technologies such as dye-sensitized solar cells\cite{hagfeldt2010dye, kalyanasundaram1998applications}. Their strong and broad absorption in the UV and visible spectra\cite{kalyanasundaram1998applications}, combined with potential water solubility\cite{bella2015aqueous}, make them attractive candidates for the development of photovoltaic materials with low toxicity levels. 
Another key area of research is the substitution of rare metals, which are often expensive and highly toxic, by earth-abundant alternatives. There is thus a strong interest in the computational design of TMCs and the ML-accelerated exploration of their chemical space\cite{Nandy9927, Dalmau1}, aiming at limiting the costs of experimental studies to the verification of hits found \textit{in silico}.

Variational autoencoders (VAEs) have found significant success in the inverse design of small organic molecules\cite{khater2025generative} and have subsequently been adapted to TMCs. VAEs use a deep neural network (\emph{encoder}) to learn a reduced-dimensionality representation (\emph{latent space}) that is subsequently decoded by another neural network (\emph{decoder}), reconstructing the input. VAEs can be easily conditioned on target properties by steering the generation towards specific regions of chemical space via gradient-based optimization\cite{Fallani24}. 
Most VAE models have hitherto focused on the generation of ligands rather than complete TMCs, thereby avoiding additional complexity associated with the treatment of coordination numbers and geometries\cite{Lee1095}, to enable, for example, the generation of palladium catalysts for cross-coupling reactions\cite{Schilter728}. In these approaches, ligands are conveniently represented as SMILES strings\cite{weininger1988smiles}, facilitating implementation into existing VAE architectures for small organic molecules, such as the junction tree VAE (JT-VAE)\cite{jin2018junction}. 
However, since coordination sites and binding modes are not explicitly encoded, the metal-binding moieties of the generated ligands are ambiguous. This limitation was addressed by Strandgaard \etal, who introduced placeholder atoms into the SMILES representation of the ligands\cite{Strandgaard2294}. 
A more holistic framework for the unconditional and conditional generation of TMCs catalysts was later introduced with the CatDRX platform\cite{Kengkanna2399}.
Despite significant progress, the lack of explicit representations of coordination geometry remains a key limitation of VAE models for the generative design of TMCs, thereby limiting the diversity of the generated output. In this regard, the development of string representations with the widest possible support for whole TMCs is a promising direction\cite{Rasmussen63}.

Similar to VAEs, diffusion models (DMs) have been adopted for the generative design of TMCs after finding extensive application in purely organic chemistry\cite{Hoogeboom8867, Alakhdar7238}. By treating molecular geometries as 3D point clouds that are internally represented as graphs, DMs noise and de-noise this representation in a largely \textit{ab initio} manner through a learned process that can also be conditioned to enable inverse design. DMs are able to capture intricate geometric relationships, including coordination geometry and ligand denticity in TMCs, reflecting the complexity and nuances associated with $d$-orbital bonding. Pioneering work on DMs for TMCs introduced representations that, in addition to position and atomic number, also encoded membership to a certain ligand\cite{Jin4377, Cornet1793}. While requiring the \textit{a priori} specification of the number of ligands and constituent atoms, this constraint provides an additional inductive bias that guides the model towards realistic coordination environments, avoiding the generation of unphysical structures. This approach also enables fixing specific ligands to facilitate the nosing/de-noising process, while conditioning the generation of individual ligands to the remainder of the TMC. When applied to the chemical space surrounding the Vaska's complex\cite{Friederich4584}, this approach proved its ability to direct the generation towards regions of interest\cite{cornet2024equivariant, cornet20250630}. Further research combining different coordination geometries with an explicit encoding of ligand denticity enabled the generation more diverse TMCs\cite{Jin8367}.

Despite their conceptual differences from VAEs, DMs, and other deep learning methods, genetic algorithms (GAs) have been extensively used to optimize TMCs. 
In this framework, TMCs can be naturally represented in terms of their metal centre and ligand fragments (\emph{genes}), which can be evolved towards optimal properties (\emph{fitness}) through change (\emph{mutation}) and exchange (\emph{crossover}) operations. 
GAs can operate at the whole-TMC level using pre-defined ligand libraries\cite{Nandy13973, Nandy8243, pita2025evolving, Janet1064, Kneiding263, Kim202500316}, or evolve ligands from smaller fragments via string or graph representations\cite{jensen2019graph}, enabling broader chemical space exploration\cite{strandgaard2023genetic, Strandgaard10638, seumer2025beyond, Laplazae202100107, Chu8885, Frangoulis3808}, while optionally incorporating synthetic accessibility constraints\cite{strandgaard2023genetic, Strandgaard10638, seumer2025beyond}.
In practice, application-oriented TMC design, \eg in catalysis, typically requires simultaneous optimization of multiple properties, such as catalytic activity, selectivity, and stability under operating conditions. GAs address these multiobjective problems either through composite fitness functions requiring \textit{a priori} weights or through Pareto-based approaches, such as the PL-MOGA framework\cite{Kneiding263,pita2025evolving}, which identify optimal trade-offs without predefined weighting.
Compared to deep learning methods that require large amounts of data upfront, GAs collect data on-the-fly over multiple iterations (\emph{generations}) through the evaluation of the fitness function. However, the fitness can be computationally expensive requiring geometry assembly from fragments\cite{Ioannidis2016molSimplify, taylor2023architector}, geometry optimizations, single-point properties of semi-empirical or DFT quality, and conformational search\cite{strandgaard2023genetic, Strandgaard10638, seumer2025beyond}.
Several studies have explored ML-based surrogate models for rapid fitness evaluation\cite{Janet1064,Nandy13973,Nandy8243} and pre-screening of candidates\cite{Frangoulis3808}, but synergies between deep and evolutionary learning\cite{Kneiding15522} remain underexplored despite promising advances such as genetic operations in latent space\cite{Grantham14} and LLM-integrated GAs\cite{Lu32377}.

\textbf{Open challenges.} Overall, there is active research on generative AI for the inverse design of TMCs encompassing both deep learning and evolutionary frameworks. These efforts will make a sustainable impact both by reducing the cost of expensive electronic-structure calculations for these molecules while also enabling their application in green chemistry. Most applied studies to date have leveraged GAs, for example for the design of catalysts and chromophores, whereas the deep learning applications have focused more on methodological aspects. While many methods try to maximize data efficiency and reduce computational cost, the use of pre-training and data-augmentation strategies, as well as the integration of ML into evolutionary frameworks, remains limited highlighting avenues to increase their sustainability. A further key challenge is the scarcity of generally accepted benchmarks that would enable direct, quantitative comparison of different methods, thereby facilitating the definition of future directions to advance the field.

\subsubsection{Generative AI for inorganic materials}

Inverse design has been very successful in expanding the vast space of known inorganic crystals, proposing new materials such as solar panels \cite{choubisaCrystalSiteFeature2020,zhaoHighThroughputDiscoveryNovel2021}, catalysts \cite{kimInverseDesignPorous2020a, zhouPracticalInverseDesign2026}, and low-supply-chain-risk materials \cite{zeniGenerativeModelInorganic2025}. The functionality of these models has been the focus of several recent reviews \cite{liMaterialsGenerationEra2025, Cheng2026, metni_generative_2025, wangLeveragingGenerativeModels2025}.
While common generative models for crystal structure design have the same architectures as their molecular counterparts (VAEs, GANs, LLMs, DMs, or flow models (FMs)), the structural differences between crystals and molecules require a few extra considerations regarding the representation. Next to the invertibility, invariance, rotational, and translational symmetry required for representing molecules, the representation of crystal structures is complicated by periodicity, symmetry, and lattice vectors and angles. Ideally, a crystal's invariant representation includes these conditions, while still being of a manageable size so the representation can be used in the training process of a model\cite{wangLeveragingGenerativeModels2025, court3DInorganicCrystal2020a}.

Recently, the importance of invariant descriptions yielding a unique representation for any given material has been at the center of critics: in 2023, Cheetham and Seshadri questioned the novelty of many materials proposed in Google's GNoME model\cite{cheethamArtificialIntelligenceDriving2024}; in 2024, analyses suggested that A-lab's novel structures largely consisted of disordered versions of known compounds\cite{leeman_challenges_2024}, and, most recently, when data leakage was found in Microsoft's MatterGen, showing that structures which had been claimed to be novel were in fact part of the training set\cite{juelsholtContinuedChallengesHighThroughput2025}. Finding an invariant, fully invertible representation for generative AI for crystalline inorganic materials thus remains an unsolved challenge \cite{renInvertibleCrystallographicRepresentation2022a, wangLeveragingGenerativeModels2025}.

Currently, there are many different ways to represent a crystal as input for a generative model, each having its own advantages and disadvantages. While approaches analogous to molecular representations based on strings\cite{xiaoInvertibleInvariantCrystal2023, wangSLICESPLUSCrystalRepresentation2025} or graphs \cite{xieCrystalGraphConvolutional2018, isayevUniversalFragmentDescriptors2017} exist, we focus here on three approaches for crystal structures.
One main approach is based on density-based representations built with voxels, which can easily be made periodic via convolutions. A prominent example is iMatGen\cite{nohInverseDesignSolidState2019}, which encodes atomic positions and cell vectors into a set of 2-D images; other examples are CGCNN\cite{court3DInorganicCrystal2020a}, CCDCGAN\cite{longConstrainedCrystalsDeep2021}, and ZeoGAN\cite{kimInverseDesignPorous2020a}. However, the size of the representation requires large-memory data pre- and post-processing to encode and extract atomic positions, limiting its application to a few elements and certain materials systems\cite{yanStructurePredictionMaterials2023}.

An alternative approach are symmetry-based representations, which often directly embed elements, space groups, or Wyckoff positions under the assumption that atoms sit on distinct sites in the crystal lattice. These lean representations allows for a coverage of large crystal spaces as demonstrated in CubicGAN\cite{zhaoHighThroughputDiscoveryNovel2021}, and it is possible to tweak the embedding or point group/crystal site relations such as in PGCGM\cite{zhaoPhysicsGuidedDeep2023}, WyFormer\cite{kazeevWyckoffTransformerGeneration2025}, and SymmCD \cite{levySymmCDSymmetryPreservingCrystal2025}. Symmetry-based representation offer high sample efficiency and validity, with invariance limited only by atom permutations and size. The representation is limited to ordered structures, which, however, can also be a desired design feature.

More flexibility is provided by matrix representations, where atoms are represented as ordered point clouds, such as in the Crystallographic Information File (CIF) format. This inversion-free, precise, and data-efficient approach has been introduced by the Composition-Conditioned Crystal (CCC-)GAN\cite{kimGenerativeAdversarialNetworks2020a}, and has since been expanded to include further material properties such as the reciprocal-space Fourier-transformed properties (FTCP\cite{renInvertibleCrystallographicRepresentation2022a}).
Microsoft's recent diffusion model MatterGen\cite{zeniGenerativeModelInorganic2025} demonstrates the strength of this flexible representation, with generative tasks covering a large composition and symmetry space. Matrix representations lack invariance, which is often acquired by data augmentation. As a result, generated structures need careful consideration.

Inverse design with generative models already works well in lower data regimes. However, the models thrive on larger datasets, which allow for a higher space coverage and often higher precision. Nevertheless, even at scale, the Pareto front between diversity and fidelity restricts the learning of the underlying data distribution\cite{turkAssessingDeepGenerative}. Thus, as accuracy limits persist, research focuses on enhancing and developing crystal representations and models, as well as strategies to effectively generate and use data, since generative models, as all machine learning models, heavily rely on the quality and quantity of training data.
A large amount of training data exists; see, \emph{e.g.}, Metni \emph{et al.} for a summary\cite{metni_generative_2025}, and yet, many datasets are biased on compositions and space groups, lacking also labels for design tasks of interest. Nevertheless, directly using or building upon openly available databases can significantly reduce the computational effort necessary to explore the design space of interest.
Sometimes it is sufficient to expand existing data by augmentation, \emph{e.g.}, adding rotated unit cells to the dataset, as in CCC-GAN\cite{kimGenerativeAdversarialNetworks2020a} or exploiting symmetries by permutation\cite{turkAssessingDeepGenerative}, to employ over- and under-sampling strategies to balance biases or to use transfer learning or fine-tuning \cite{zeniGenerativeModelInorganic2025}.

In case new data needs to be generated, smart methods that maximize information gained from experiments such as design of experiments (DoE)\cite{boxStatisticsExperimenters1978} or incremental online learning schemes that integrate a space exploration tool such as reinforcement learning are desirable\cite{ueharaFeedbackEfficientOnline}, \eg as MatInvent\cite{chenAcceleratingInverseMaterials2025}. Another promising direction are uncertainty-aware models, which provide prediction confidence that can flag structures in high-uncertainty regions as well as differentiable models, that provide informative gradients which can stabilize training\cite{wangLeveragingGenerativeModels2025, chenAcceleratingInverseMaterials2025}.

\textbf{Open challenges.} Bridging to realistic conditions is another essential goal for practical model use and for comparison with experiments \cite{wangLeveragingGenerativeModels2025, liMaterialsGenerationEra2025}. Much underlying data derives from DFT at 0~K with limited cell sizes and known approximation errors. Translating such predictions into applications often requires modeling temperature and pressure effects, large supercells, or advanced representations for complex stoichiometries and defect chemistries, and an explicit treatment of disorder and amorphous structures. Predictions based on perfect crystals miss these property-relevant structural features, complicating comparisons with experiment and even stability assessment, which is commonly done by referring to the formation energy or the zero-temperature convex hull \cite{zeniGenerativeModelInorganic2025, metni_generative_2025}. Including these effects is an emerging field: the first generative models for disordered inorganic crystals and classifiers for substitutional disorder begin to address these gaps\cite{petersen_dis-gen_2025, jakob_learning_2025}, tackling the unresolved issues around disorder prediction, structural uniqueness, and dataset validation.

Next to the success rate and realistic conditions, many generative models are also limited in the amount of possible property constraints\cite{zeniGenerativeModelInorganic2025}. In general, it is desirable to optimize on multiple-objectives to penalize scarce/critical/toxic elements, prioritizing earth-abundance, low-supply risk, and recyclability. Many models already allow for such multi-objective prediction\cite{yeConCDVAEMethodConditional2024, chenAcceleratingInverseMaterials2025, boonkirdAreQuantumMaterials2025}, though balancing the conditions during training still poses challenges. Overly strict constraints can miss breakthrough chemistries, especially since many unexploited parameters, such as composition, space group, defects, disorder, grain boundaries, entropy, and environmental conditions, can heavily influence the material's properties while being concealed from the model. 

Another desirable condition is the incorporation of synthesizability information (\emph{e.g.}, precursor availability, thermodynamic feasibility) to reduce dead-end attempts or ideally even the accurate prediction of synthesis paths \cite{liMaterialsGenerationEra2025}. Despite algorithmic advances, translating generative inverse-designed crystal structures into actual experimental synthesis is yet not fully demonstrated\cite{renInvertibleCrystallographicRepresentation2022a}. Integrating automated, AI-driven laboratories can provide rapid feedback on proposed designs, which allows autonomous experimental validation and iterative improvement of inverse design models\cite{sanchez-lengelingInverseMolecularDesign2018}. Hence, closing the loop from design to synthesis and validation is an essential step for sustainable material generation.

\subsection{Functional roles of language models in scientific discovery}

\subsubsection{Language models as predictive tools}

\begin{figure}[t!]
    \centering
    \includegraphics[width=0.7\linewidth]{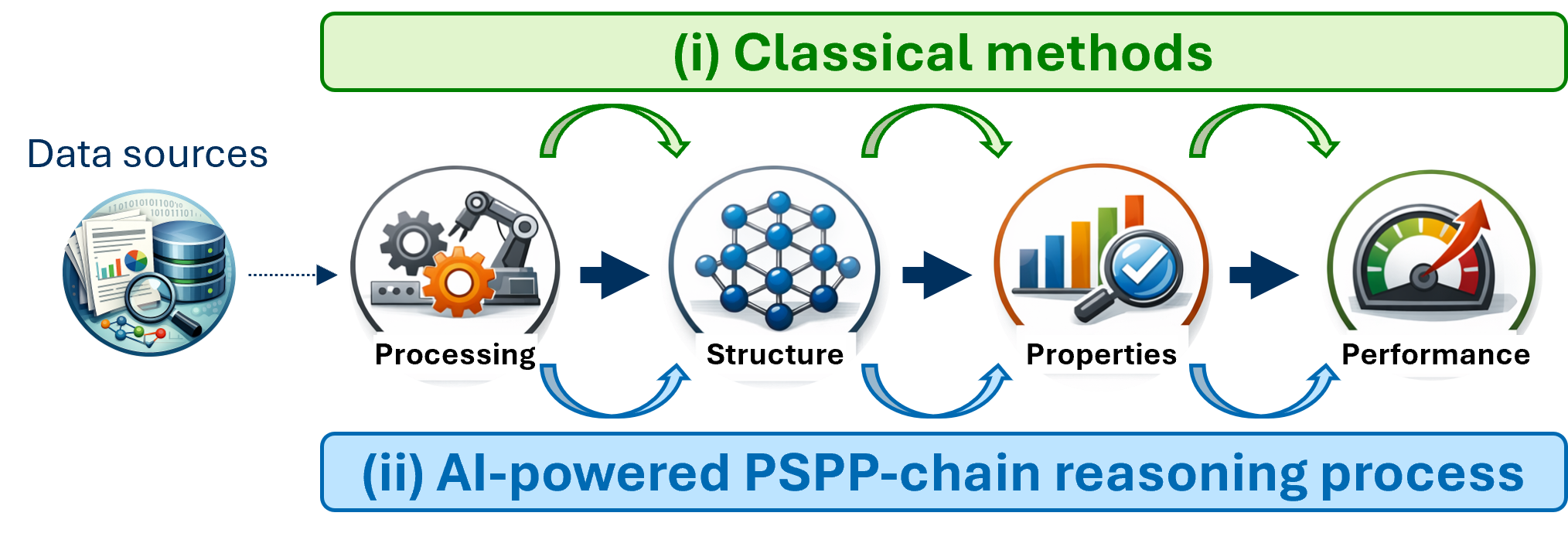}
    \caption{\textbf{Classical and AI-powered approaches along the PSPP relationship \cite{Ehrenhofer2025LLM_paper}}: Classical approaches (i) to connect the different parts range from quantum theory and molecular dynamics approaches up to continuum models. The AI-powered PSPP-chain reasoning process (ii) offers an alternative approach, where every connection is represented by a data-driven model. Models can also leapfrog steps, such as Processing-Property models \cite{Wang2024prediction_hydrogel}.}
    \label{fig:image_llm_pspp}
\end{figure}

One way to include \textit{world-knowledge} into materials design is by leveraging today's Large Language Models (LLMs) as connections in the Processing-Structure-Property-Performance (PSPP) relationship\cite{Ehrenhofer2025LLM_paper,Ehrenhofer2025smart_materials_informatics}, see Fig. \ref{fig:image_llm_pspp}. The main approach behind this kind of model is the representation of the meaning of a (sub-)word in an abstract and multi-dimensional embedding space. The similarity of words or the connection of concepts are represented by distances between these embeddings. Token-based LLMs that are based on the generative pre-trained transformers (GPT) architecture \cite{Vaswani2017} are extensively used in today's economy and start to reshape various fields, from science to society.

There are various approaches to integrate LLMs into the material development process \cite{zhang2025large,zhang2025scientific}. While the focus was previously on extractive language models with capabilities to analyze data (see next section), approaches with generative models such as ChemGPT \cite{frey2023neural} or ChemLLM \cite{zhang2024chemllm} are mostly focused on providing a conversation interface with a material-specialized chat model. The current section aims at highlighting paths towards deep integration of LLMs in the design workflow at the interface between materials science and engineering design. 

The starting point is, as characteristic for data-based approaches, the data. Large general-purpose (i.e., non-specialized with respect to materials science) datasets are used to train these models; furthermore, additional steps, such as reinforcement learning with human feedback, are used to enhance their capabilities for generating \textit{useful} output. Please note that \textit{useful} in this context is subject to various contrary goals in the field of tension between user retention and scientific accuracy.

It is conceivable that the similarity of materials, the concepts of production/fabrication/processing, the meaning of material parameters, and the definition of performance characteristics can be represented in the same way. If this is accurate, LLMs can be applied in a PSPP-chain reasoning process that predicts the performance of a specific realization of a material in a specific context \cite{Ehrenhofer2025smart_materials_informatics}. In previous work, we have shown that LLMs contain \textit{material knowledge} on the level of the Periodic Table of Elements \cite{Ehrenhofer2025LLM_paper}. A systematic approach to the representation of LLM-intrinsic knowledge, as proposed, for example, in the fields of finance and history \cite{Giordano2026}, can be highly beneficial for the development of such PSPP-chain reasoning models.

\textbf{Open challenges.} In the context of \textit{sustainably} applying ML approaches for material discovery, it is questionable if large, general-purpose models, which include a variety of scientific and non-scientific sources, are the right tools: It is yet to be proven that the data points from such diverse fields that are covered by huge datasets such as the Pile \cite{Gao2020a} or the Common Crawl \cite{Patel2020} lead to a generalization that is beneficial for the field of material science. Therefore, we propose applying specialized \textit{open} small language models and rigorously benchmarking them for material knowledge. The LLM benchmarking tool by the authors is a first small step in establishing this kind of materials-centered testing framework; this can lead to an increase in material knowledge of models because the presence of benchmarking tools usually leads to the inclusion of additional aspects in LLM development. The \textit{openness} and size of the models are another crucial aspect in the context of scientifically applicable LLMs: Only the knowledge of datasets, training parameters, as well as weights, will allow the construction of models with high scientific impact. In our opinion, \textit{sustainably} integrating generative AI models into material development workflows means using small and specialized models with high material knowledge, instead of defaulting to non-specialized solutions like ChatGPT. In this way, the application of generative AI will help in increasing research efficiency for material development, reducing the environmental impact of extensive testing, while keeping the additional costs of computation low. 

Last, but not least, generative models such as large or small language models can be used to extract data from unstructured text using processes such as Retrieval-Augmented-Generation (RAG). In this approach, text from primary sources (e.g., scientific papers) is separated into chunks (e.g., sentences, word-groups, or chapters) and the combined embedding of the text inside the chunk is generated based on the tokenizer/vocabulary, attention mechanism, positional encoding layer, etc., of an LLM. Then, large unstructured datasets, such as a collection of scientific papers, can be added to the context of a generation, and specific extraction questions can be answered by the LLM. Various specialized embedding models can be found, e.g., in the leaderboards at huggingface (https://huggingface.co/spaces/mteb/leaderboard). However, it must be noted that LLMs based on generative pre-trained transformers are not the ideal model architecture for this task; instead, extractive models built on the bidirectional encoder representations from transformers (BERT) architecture are much better suited for this task, as will be shown in the next section. 

\begin{figure}[t!]
    \centering
    \includegraphics[width=0.75\linewidth]{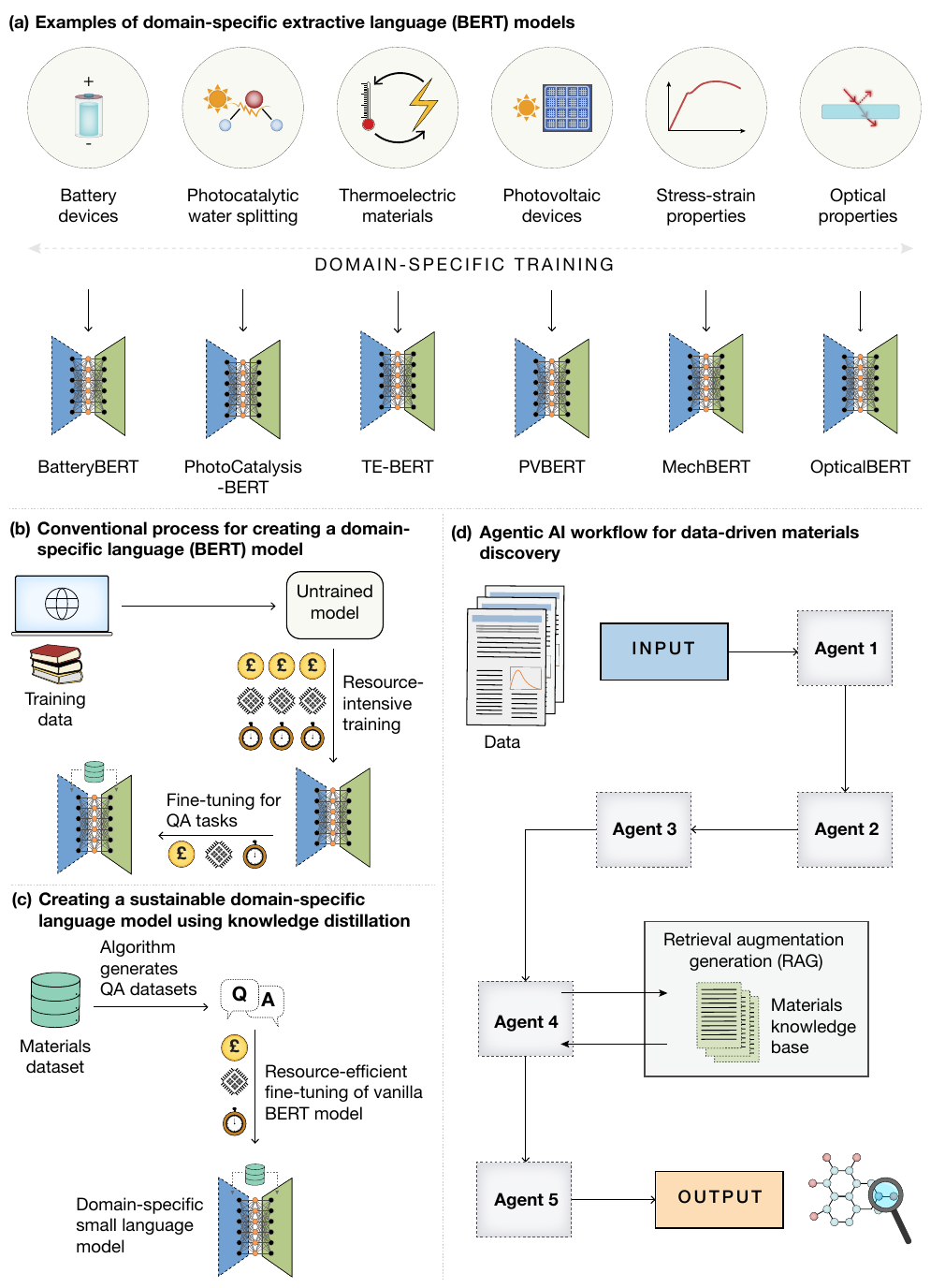}
    \caption{\textbf{Conceptual overview of sustainable extractive language models for domain-specific applications in materials science:} (a) examples of published domain-specific BERT models derived from structured datasets across different domains (e.g., batteries, photocatalysis, thermoelectrics, photovoltaics, and stress–strain properties); (b) conventional training process for a domain-specific language (BERT) model; (c) resource-efficient creation of domain-specific models via knowledge distillation; (d) integration of small, domain-specific BERT models as agents within an agentic AI workflow for data-driven materials discovery.}
    \label{fig:extractivemodels}
\end{figure}

\subsubsection{Language models as extractive tools}

Extractive language models stand to offer higher-quality data retrieval than generative language models. In the case of those based on the BERT architecture, this is because its bidirectional nature allows data representations of contextualized language to be learnt\cite{Vaswani2017}. Therein, its self-attention heads provide a means to decipher relationships between tokens (\eg words, sub-words) from a given input context. Once suitably trained on a large corpus, a BERT model will hold a fully contextualized representation of this entire text. 

Chemical language is unique and markedly distinct from natural languages (\eg English); therefore, this fine-grained contextualization becomes particularly important for chemical applications. These niche applications tend to be governed by chemical structure-property relationships, which are, in turn, only tractable if one can concurrently extract correlated data about chemical properties and their chemical names. Moreover, chemical names can take many forms of identity (\eg IUPAC name, InChI, SMILES,\cite{smiles} SELFIES\cite{selfies}); so, one needs to solve the complex problem of chemical-named-entity recognition while one simultaneously extracts correlated data\cite{elsa}. This is not trivial; indeed, such a task is somewhat removed from a simple value-extraction task that can now be readily carried out by many types of language models. 
A trained BERT model can be fine-tuned for question-answering tasks such as launching a query about the property of a particular material; as such, it can field textual queries from users in the style of prompt engineering.

BERT models are open-source, which holds many benefits, including the opportunity to train custom BERT models with bespoke text so that they suit a given domain-specific application. In the field of materials science, MatBERT\cite{matbert}, MatSciBERT\cite{matscibert} and BatteryBERT\cite{batterybert} models were the first materials-focused BERT models to be published. MatBERT and MatSciBERT models were trained to represent broad knowledge about all of materials science. 
The BatteryBERT model was the first published language model whose application was geared to a specific domain of materials science: the field of battery-device materials. This application used state-of-the-art technology at the time when the BatteryBERT paper was published; although many new capabilities, such as retrieval augmented generation (RAG) offer a far wider range of applications\cite{rag1,rag2,rag3}.
The higher performance of BatteryBERT model compared to general-purpose BERT models led to the development of a wide range of domain-specific BERT models for given fields of materials science (\eg Fig. ~\ref{fig:extractivemodels}a). These fields include optics (OpticalBERT)\cite{opticalbert}, thermoelectrics (TE-BERT)\cite{tebert}, optoelectronics (OE-BERT)\cite{oebert}, stress-strain engineering (MechBERT)\cite{mechbert}, photovoltaics (PVBERT) and its sub-domains that befit perovskite solar cells (PSCBERT) and dye-sensitized solar cells (DSCBERT)\cite{pvbert}, photocatalysis for water splitting (PhotocatalysisBERT), and the entire area of physical sciences (PhysicalSciencesBERT)\cite{photocatalysisbert}. The entire collection of BERT models is available as a dedicated module on the Digital Materials Foundry that is hosted by the Henry Royce Institute (https://www.royce.ac.uk/programmes/digital-materials-foundry/).

BERT baseline models were further tuned with SQaAD2.0 \cite{squad2} to generate a BERT-based question-answering (QA) tool for information extraction of materials properties from text \cite{sippila_2025}. To evaluate model performance, it was necessary to build manually annotated ground-truth datasets such as PV600, which contains 600 text snippets on perovskite materials bandgaps \cite{pv600}. Comparative performance across five different BERT-based QA tools indicated that awareness of materials science domain knowledge is a considerable advantage in extraction tasks, with BERT underperforming, while MatSciBERT and MatBERT excelled with QA. These models were next tested against LLMs on extraction tasks, with both paid (OpenAI’s GPT-4o \cite{gpt4o}) and free (Llama and Mixtral) model types. Here, MatSciBERT-QA performed competitively close to the best GPT-4o model, and (at the time), better than open LLM models \cite{pv600}. All the above BERT-based QA models were encoded in a web application Materials Property Information Extractor (M-PIE) \cite{mpie} to facilitate easy testing and information extraction for non-experts. These findings demonstrate of the good capability of BERT models compared to LLMs in extractive applications, despite their more modest size and parameters. 

These BERT models are naturally much smaller than general-purpose language models, both in terms of their internal parameters that are governed by the BERT architecture and their data, where the focus is on a high-quality, domain-specific corpus whose specialized knowledge is implicitly limited. 
Indeed, given the many technical advances in language models since the originally styled BERT models were first reported, BERT models would probably be best classified as small language models (SLMs) these days. 
Nonetheless, the compute needs for training BERT models are still sufficiently high that supercomputing facilities are necessary if they are to be created in a timely fashion (see Fig. ~\ref{fig:extractivemodels}b). In addition, the original BERT model, that was naturally trained from scratch, incurred a very considerable level of carbon emissions during its training process\cite{co2bert}. None of this bodes well for achieving the sustainable ML goals of this perspective.

Fortunately, recent work has found a way to overcome the need to train domain-specific BERT models\cite{pvbert,tebert}. The method uses a form of knowledge distillation that is illustrated in Fig. ~\ref{fig:extractivemodels}c. The conceptual idea is to keep the domain-specific corpus outside the language model, but use it to create a vast number (100,000s) of pairs of domain-specific questions and answers that are then employed to simply fine-tune a vanilla BERT model; fine-tuning a language model is computationally very modest, in stark contrast to fully training a language model\cite{co2bert}, as also discussed for MLIPs (see Sec. \ref{sec:MLIPs}). The performance of domain-specific language models obtained with this approach has been shown to be comparable to that of conventionally trained models. Thus, this form of domain-specific knowledge distillation offers substantial computational savings without significantly compromising the performance of the resulting model.
This result conceptually stands to reason when one considers that the knowledge-based data (from the corpus) have been distilled into the domain-specific question-answering dataset while the language model is primarily used for its operational architecture. This also means that the performance of a language model created by this method is not dependent on its size, unlike the case of conventionally trained language models. Rather, the size of the question-answering dataset used for fine-tuning the domain-specific language model will influence its performance.

Notwithstanding the benefits of this method, its viability is contingent on an auxiliary set of algorithmic steps that re-frame the domain-specific corpus (source data) into the domain-specific question-answering dataset (input for fine-tuning the language model). The method adopted to date\cite{pvbert,tebert,nbert} is a three-step process that first auto-generates a materials database from the domain-specific corpus, using the chemistry-aware text-mining tool, ChemDataExtractor\cite{cde1,cde2,cde21}. Each data record of the resulting database will contain the chemical name, one or more properties and the document object identifier (DOI) of the original paper containing the context from which the record was extracted.
The second step of this process involves locating the relevant sentence from each of the papers that contains extracted chemical and property data residing in each record.
The third step algorithmically reframes this sentence into a set of conditional questions whose answers equate to the {chemical, property} records of the materials database. 
Thus, a materials database can be reframed as a large question-answering dataset and used in a prompt-engineering fashion to fine-tune a vanilla language model, producing a high-quality domain-specific language model at low computational cost. 

\textbf{Open challenges.} Small and specialized language models, such as materials-domain-specific BERT models, are valuable in their own right; although, they stand to have markedly added value if they are employed as AI agents within a goal-oriented agentic AI workflow. Thereby, they could be incorporated into an operational workflow that links them together with other AI agents in an integrated fashion; in a way that they can work in concert to achieve a complex goal whose sum-is-greater-than-the-parts of all AI agents when considered individually. An agentic AI workflow that tracks a carefully assembled sequence of task-driven models or tools to serve such a goal is essential for success.
For example, a materials-domain-specific BERT model could be incorporated into an agentic AI pipeline, whose goal is to realize data-driven materials discovery (see Fig. ~\ref{fig:extractivemodels}d). In such a scenario, the extractive nature of this BERT model might be used, in concert with a RAG, to act as an AI agent that retrieves and faithfully reproduces a set of materials-domain-specific data or extract of text that can be screened by another type of AI agent; or the BERT model could be deployed as an AI agent that pre-processes or validates a key part of a ‘design-to-device’ pipeline for materials discovery\cite{coledesigntodevice}. 
The highly focused and automated nature of many AI agents tends to furnish them with a highly energy-efficient utility. In turn, their corresponding agentic AI workflows tend to be manageable in terms of achieving sustainable ML requirements.

\section{Optimizing sampling for sustainable exploration of chemical spaces}

\subsection{Active learning and uncertainty-driven frameworks}

\subsubsection{Uncertainty-aware exploration of energy landscapes}

Training of accurate and generalizable ML models requires high-quality datasets encompassing a vast chemical space of molecules or materials. In the case of ML interatomic potentials (MLIPs), these datasets further incorporate out-of-equilibrium, high-energy structures that are necessary for stable and robust molecular dynamics (MD) simulations. The generation and labeling of such datasets requires a significant investment of computational resources, often involving the manual generation and selection of thousands of structures of varying composition, which must be labeled using reference quantities obtained from electronic-structure theory. While the cost of this procedure can be mitigated by reducing the level of theory used for data labeling or by employing data-efficient ML architectures, such as equivariant MLFFs discussed in Sec. \ref{sec:MLIPs}, the preparation of training data remains a labor-intensive process.

Active learning (AL) approaches aim to automate and streamline the data selection to provide accurate and robust MLIPs\cite{settles_active_learning,survey_al_materials_2026}. AL selects training data based on the current state of the model, prioritizing data points for which predictions are less reliable or prone to extrapolation. This process reduces the epistemic uncertainty of the model, that is, uncertainty arising from model error, thereby improving performance while reducing redundancy in the newly selected data. AL thus significantly improves the data efficiency of the data selection process compared to manual or random sampling strategies.
A typical AL workflow consists of training a surrogate ML model on an initial version of the training dataset. This model is then used to generate a new set of structures, for example, through MD simulations, which are subsequently used to iteratively improve the model via active selection of data points based on a predefined criterion. The selected data are then labelled using reference electronic-structure calculations, added to the training dataset, and the model is retrained. In the field of MLIPs, commonly used AL approaches are based on structural similarity and uncertainty quantification. The uncertainty quantification in ML and AL approaches was recently discussed in several reviews\cite{uncertainty_in_atomistic_modelling,doi:10.1021/acs.chemrev.4c00572,DaiAdhikariWen+2025+333+357}. Here, we list the representative approaches used in AL for MLIPs. 

Among methods that evaluate structural similarity within datasets, common approaches either assess configurations directly \cite{Sivaraman2020} or identify novelty in descriptor space\cite{D_optimality}, such as that defined by Smooth Overlap of Atomic Positions (SOAP) \cite{zhang_explicit_solvent} and Atomic Cluster Expansion (ACE) descriptors\cite{AL_for_ACE}. These approaches label data points that lie outside the span of the existing training data, signaling potential extrapolation and the need for additional reference calculations.

Natural framework for AL is provided by Gaussian process regression (GPR) models, as they output predictive uncertainty in the form of a mean and variance\cite{Guan18042018}. This information can be leveraged for on-the-fly selection of data points during MD simulations, triggering reference calculations and model retraining when the uncertainty exceeds a predefined threshold\cite{PhysRevB.100.014105, kozinsky_interpretable_bayesian}. In contrast to GPR, neural network-based potentials (NNPs) do not natively provide uncertainty estimates, as their predictions consist solely of point values. In this case, model uncertainty is commonly approximated using query-by-committee approaches\cite{query-by-committee}. Here, several models are trained in parallel using, for example, different subsets of the training data or different initial weights. During active learning, new data points are selected based on the variance of the committee’s predictions\cite{behler_tutorial_review,roitberg_AL,schran_committee,PhysRevMaterials.3.023804,ceriotti_uncertainty_MD_2021}. The variant of the ensemble approach, the deep ensemble framework, adds an output function to the NN estimating the variance of the predicted quantity\cite{deep_ensembles,madsen_deep_ensembles}. To reduce its substantial computational cost, the sampling of so-called shallow ensembles, where only the last NN layer is varied, was introduced as an efficient alternative\cite{ceriotti_shallow_ensembles}.
The cost associated with training model ensembles further motivated the development of single-model uncertainty quantification schemes. For instance, uncertainty can be estimated using dropout NNs, in which outgoing connections from randomly selected nodes are eliminated during inference\cite{dropout_2016,dropout_Wen2020}. Another approach employs a Gaussian mixture model trained on NN features to compute force uncertainties for individual atoms in the system\cite{Gaussian_mixture_model}. 

\textbf{Open challenges.} Recent advances in MLIPs, driven by new architectures and large, diverse datasets, have enabled the development of general-purpose models (often termed foundation or universal models). These models can generalize across broad chemical spaces and provide stable out-of-the-box performance for many systems\cite{batatia_foundation_2025,kabylda2025molecular}. However, specialized tasks, such as chemical reactivity modeling or rare-event sampling, typically require additional fine-tuning to achieve high accuracy. In this setting, AL approaches offer a principled strategy for constructing representative, data-efficient fine-tuning datasets. Nevertheless, the integration of general-purpose MLIPs into conventional AL workflows remains insufficiently explored. For instance, recent studies have proposed uncertainty-estimation techniques for pre-trained models, including readout ensembling and quantile regression \cite{uq_foundation_models}, as well as adaptations of query-by-committee schemes through multi-head committee architectures \cite{schran_multihead_committees}. These developments highlight both the promise and the methodological challenges of extending AL concepts to next-generation MLIPs.

\subsubsection{Sustainable materials sampling with costly simulations}

While MLIP development has experienced transformative advances, simulations of many complex materials such as heterostructures, hybrid organic/inorganic materials, magnets, as well as surface and interface studies, still require \textit{ab initio} approaches. Here, DFT delivers a favorable balance between QM accuracy and computational cost. However, complex chemistries, heavy basis sets and large unit cells simulations render extensive sampling computationally expensive, and this makes exploration of chemical and property spaces prohibitive. This issue can be partly mitigated by deploying high-throughput workflows on very high-performance computing resources but that approach is not always sustainable from the viewpoint of energy and resources invested. 

Here, probabilistic AL using Bayesian optimization offers an attractive alternative with its smart sampling approach. 
Such a probabilistic approach was encoded into the Bayesian Optimization Structure Search (BOSS) Python framework for low-throughput materials optimisation \cite{todorovic}. BOSS generates N‑dimensional surrogate models for materials' energy or property landscapes, where it is easy to identify global and local optima. 
Moreover, such models are chemically interpretable, and allow to inspect the variation in materials properties across the search space of experimental parameters. The models are iteratively improved until convergence by sequential or batch input of new simulations, selected by acquisition strategies. This results in lean yet highly informative computational datasets, and allows to identify optima with modest computational sampling.

In simulations, BOSS was applied to gain insight into both organic and inorganic materials, as well as their interfaces. Studies of kinetic disorder in perovskites \cite{jingrui}, spin interactions in magnets \cite{cesare} and solid-solid interfaces \cite{solid-interface} improved the understanding of structure-property relationships in complex solids. Hybrid organic/inorganic interfaces were tackled by studies of ligand-protected clusters \cite{}, molecular adsorbates \cite{camphor-afm} and thin film growth \cite{egger}, and the same tools were deployed to explore gas sensors \cite{llobet} and biosensors \cite{estradiol}. 
When big data ML is required, AL selection strategies can help to reduce computational load by strategic labeling of materials data. Strategies that combine both prediction uncertainty and structural diversity cut large datasets to half their size while maintaining model performance \cite{homm_reduction}, and they enable rapid discovery of materials with targeted properties from broad collections of unlabeled candidates. 

\textbf{Open challenges.} While BO is data-efficient, there is potential to further reduce sampling requirements and accelerate discovery. Poor model scaling with increasing search dimensionality could be resolved by mapping the search to a low-dimensional feature space \cite{moriconi2020high}. Categorical as well as continuous variables could be incorporated into joint mixed-variable BO approaches \cite{zhang_2025}. 
Incorporating data gradients into GPR allows to use computed forces in addition to energy data to fit energy landscapes with fewer simulations \cite{mikael_thesis}. Further research into specialized kernels for materials and acquisition functions that target multiple objectives or incorporate computational cost information could additionally accelerate sampling. Lastly, advanced multi-task GPR models allow to layer information from multiple information sources or multiple simulators. This facilitates knowledge transfer from one task to inform the sampling on the other, resulting in a faster discovery of optima \cite{mtbo_herbol_2020}. Human-in-the-loop approaches have demonstrated that informing BO sampling with human preference data leads to faster convergence towards desired solutions \cite{armi}. More community efforts are required to investigate these approaches and gain insight into best practice guidelines for maximal data-efficiency and sustainability.

\subsection{Data efficiency in machine-learned interatomic potentials}

MLIP models depend crucially on the amount and nature of their training data, and indeed data are one of the frontier challenges for methodology in this field \cite{BenMahmoud2024, Kulichenko2024}. Early MLIP models were based on carefully hand-crafted datasets: examples include the first NN potential for the phase-change memory material GeTe \cite{Sosso2012}, and iteratively constructed potentials capturing materials properties of tungsten \cite{Szlachta2014} or the structure of amorphous carbon \cite{Deringer2017}. More recently, there has been a surge of interest in general-purpose MLIP models that are trained on datasets comprising wide ranges of compositional space, leveraging the ability of modern graph-based architectures to scale favorably with the number of chemical elements involved \cite{chen2022universal, deng2023chgnet, Zhang2024, batatia_foundation_2025}. In other words, while early MLIPs for inorganic materials typically focused on single-element or binary systems, we can now describe most elements from the Periodic Table with a single model (at least in principle!).

In the context of the SusML workshop, the question of `sustainability' links to how we can train and evaluate MLIP models {\em efficiently}: with small and manageable datasets that still convey the required information, and with small and economic models for computationally demanding downstream tasks -- say, realistic-scale materials simulations with millions of atoms and more. This type of efficiency is particularly important when computational resources are limited, and when the carbon footprint of high-performance computing is taken into account.

\begin{figure}[t]
    \centering
    \includegraphics[width=0.95\linewidth]{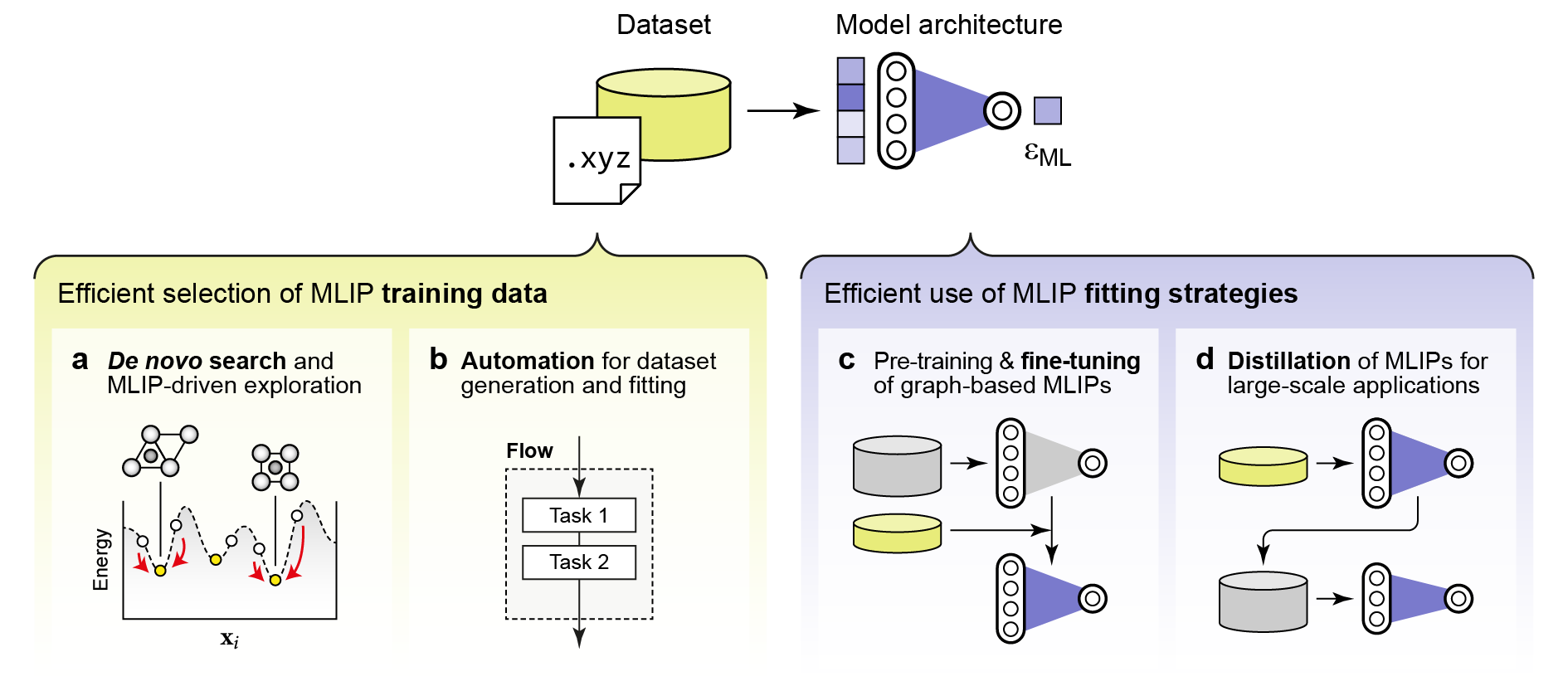}
    \caption{\textbf{Data efficiency for machine-learned interatomic potential (MLIP) models.} This schematic overview summarises two aspects, building upon a talk at the SusML workshop. 
    The first aspect ({\em yellow}) is being efficient in the selection of training data; this comprises (\textbf{a}) the exploration of configurational space with evolving MLIP models and (\textbf{b}) the use of automated workflows.
    The second aspect ({\em blue}) is being efficient in the way that the models are fitted. The pre-training of graph-based models, and their subsequent fine-tuning on much smaller amounts of data (panel \textbf{c}), is particularly timely in the age of large general-purpose (also referred to as `foundational') MLIPs. Furthermore, evaluation speed at runtime can be improved by model distillation (panel \textbf{d}): the sketch on the right illustrates a data-driven teacher--student approach \cite{Morrow2022} which uses an accurate, but slow model to create training data and fit a more specialised, much faster model to those.
    The drawing at the top is similar to one in a recent Comment article, first published in Ref. [\citenum{BenMahmoud2024}] by Springer Nature.
    Panels \textbf{a} and \textbf{b} are adapted from Ref. [\citenum{liu_automated_2025}], which is published under a CC BY licence (http://creativecommons.org/licenses/by/4.0/).}
    \label{fig:Figure_vld}
\end{figure}

There are two primary components to an MLIP model: (i) the reference dataset of atomic structures and ground-truth labels, and (ii) the model fitting architecture, which comprises the choice of representations as well as the regression task itself (see Fig.~\ref{fig:Figure_vld}). Accordingly, there are two main areas in which efficiency and sustainability are important.

The first area is the efficient selection of MLIP {\em training data}. A natural choice early on was to use structural snapshots from {\em ab initio} molecular dynamics (AIMD), driven by DFT, to explore configurational space. However, it is much more efficient to run this exploration with MLIPs if possible, even with early versions that are not yet providing a `perfect' result, and then to use the comparably expensive QM reference method only for the purpose of computing single-point energy, force, and stress labels. This way, instead of choosing multiple snapshots from a single AIMD trajectory, one can run large amounts of parallel, uncorrelated simulations and select from a wider pool of structures \cite{Deringer2017}. Alternatively, one may accelerate the AIMD runs with on-the-fly learning directly \cite{Li2015, Jinnouchi2019, Stenczel2023}, which has been used with success to create purpose-specific training datasets and targeted MLIP models \cite{Baldwin2024, ElMachachi2024, VOLKMER2026121844}. In all these cases, AL techniques can help to select representative data for labeling, as discussed in the preceding section.

The idea of iterative dataset construction has been combined with established approaches to crystal-structure prediction \cite{Deringer2018, Tong2018, Podryabinkin2019}. The process is sketched in Fig.~\ref{fig:Figure_vld}a; the drawing emphasizes how the process involves a broad coverage of structures, the ones that are relevant and also those which are not. The latter is important to sample unfavorable regions of the PES, avoiding `false minima' \cite{Behler2021, Morrow2023}, and hence the key advantage of this wide exploration is robustness rather than accuracy. The latter can follow, of course, by further extension of the training dataset once the most relevant regions are known.

Relevant too is the use of automation and workflows (Fig.~\ref{fig:Figure_vld}b): here, the term `sustainability' mainly applies to the reproducibility of results, as well as reducing the time and effort required to build and evaluate MLIP models. The advantages of systematic automation and the associated methodological and software developments are discussed in the following section. In the context of MLIPs, specific frameworks have begun to be developed for the assembly of training datasets \cite{menon_electrons_2024, liu_automated_2025} as well as downstream evaluation and benchmarking tasks \cite{Zills2025}. A recent work highlights the usefulness of automation for atomistic ML in computational practice: automated structure-searching runs were used to create an initial MLIP training dataset for arsenic, and those data were then supplemented with iterative MD exploration, enabling structural studies of the element's amorphous modification with rather modest computational resources \cite{Liu2026}.

The second main area for `sustainability' of MLIPs is the efficient use of {\em fitting strategies} (Fig.~\ref{fig:Figure_vld}c--d): given an existing dataset, how do we fit the most accurate and most economic potential to it?
In the context of NN-based MLIPs, transfer learning has been pioneered years ago: Smith {\em et al.}\ showed how an ANI model pre-trained on comparably cheap DFT data could be adapted to coupled-cluster data, re-optimising two of the four network layers \cite{Smith2019}. 
K\"a{}ser and Meuwly discussed the role of pre-training datasets and evaluated transfer-learning approaches against experimental benchmarks for malonaldehyde \cite{Kaeser2023}.
A separate approach, mentioned here only in passing, is $\Delta$-learning \cite{Ramakrishnan2015a, Bogojeski2020,equidtb}.
In the space of materials modeling, pre-training with MLIP-generated `synthetic' data was described for simple NN models of atomic energies as a test case \cite{Gardner2023}, and subsequently for graph-based MLIPs \cite{Gardner2024}. 
More recently, fine-tuning has emerged as a highly popular route in the field, supported by the availability of very large pre-trained models: the fine-tuning of general-purpose MLIPs has been demonstrated, {\em e.g.}, for ice polymorphs \cite{Kaur2025} and organic molecular crystals \cite{DellaPia2025}, where subtle energetic differences between polymorphs are important to capture accurately.
The \texttt{MatterTune} software introduced for fine-tuning general-purpose MLIPs \cite{Kong2025} is an example of purpose-built tools that are expected to accelerate progress in this field further.

Model distillation is a strategy for improving the efficiency and sustainability of ML-driven simulations, particularly given the high computational cost of graph-based MLIPs. In 2022, it was shown how an accurate, but comparably slow MLIP model can be used to create training data for a more specialized, but much faster one: in this case, the `teacher' model was a general-purpose MLIP for silicon \cite{Bartok2018}; the `student' model was specialized for amorphous silicon, $a$-Si, and its behavior under pressure; see Ref. [\citenum{Morrow2022}] and references therein. With the distilled student potential, it was possible to efficiently study defects in a million-atom model of $a$-Si \cite{Morrow2024} as well as, subsequently, the signatures of local structural order in this canonical disordered network \cite{Rosset2025}. Recently, it was argued that distillation is a particularly useful strategy for general-purpose MLIPs: the `teacher' model covers many configurations from across the Periodic Table, but it may subsequently be fine-tuned and distilled into a `student' model which is specialized for the problem at hand \cite{Wang2025a, Gardner2025Distillation}. There are other approaches to model distillation too, which focus on the architecture itself rather than the data; these are beyond the scope of the present perspective.

\textbf{Open challenges.} Many ideas exist for making MLIPs more efficient, but the implementation of these ideas and their validation under realistic application conditions are often still outstanding challenges. Some of the ideas discussed above could well be combined in the future: say, by integrating model distillation into automated workflows. Sustainably developing the next generations of MLIPs, and sustainably applying them in practice, will become increasingly more important as these methods evolve into true mainstream approaches.

\subsection{Automated high-throughput discovery pipelines}

\subsubsection{High-throughput discovery in inorganic materials}

Computational high-throughput discovery pipelines have been in use for over 20 years, with the first pioneering works dating back to 2003\cite{hautier_finding_2019}. Since then, computing times, \textit{ab initio} methods, ML methods, and workflow managers have been massively advanced. The key ideas, however, have remained the same. With these advances, standard high-throughput searches have become computationally cheaper, easier to implement, and therefore more sustainable.

Within such a high-throughput computational pipeline, crystal structure candidates are evaluated across several tiers to identify candidate materials suitable for a specific application\cite{hautier_finding_2019}. The candidate materials can then be passed on to experimentalists or experimental pipelines. At each tier, a materials property relevant to the application will be tested using a descriptor computed by \textit{ab initio} or ML methods. Thanks to advances in computational databases (\eg Materials Project\cite{jain_commentary_2013, horton_accelerated_2025}, OQMD\cite{kirklin_open_2015}, Aflow\cite{curtarolo_aflowlib.org:_2012}, NOMAD\cite{draxl_nomad_2018}, Alexandria database \cite{schmidt_machine-learning-assisted_2023, cavignac_ai-driven_2025}), some of these properties can also be looked up within a database, making the process more sustainable. Joint APIs, such as OPTIMADE, make access even easier\cite{andersen_optimade_2021, evans_developments_2024}. Also, critical raw materials can also be considered as part of such a pipeline\cite{gaultois_data-driven_2013, dahliah_high-throughput_2021}. The tiers will be sorted by required computational times (\ie simple database look-ups will be at the beginning, while demanding \textit{ab initio} computations will be at the end). 

As mentioned above, and as part or side product of high-throughput searches, several large-scale databases, such as Materials Project and Alexandria, have been built, which now serve as the starting point for universal MLIPs, including MEGNET\cite{chen_graph_2019}, M3GNET\cite{chen2022universal}, CHGNet\cite{deng2023chgnet}, and MACE-MP-0\cite{batatia_foundation_2025}. While not specifically built with the application of MLIPs in mind, they have been invaluable in building the first universal potentials. Such databases have also helped benchmark these potentials, \eg as part of Matbench Discovery\cite{riebesell_matbench_2024}. Projects that facilitate easy data reuse for training MLIPs are also beneficial\cite{vita_colabfit_2023}. Nowadays, these MLIPs are also accelerating high-throughput searches. Additionally, a part of the code infrastructure for MLIPs has been simplified, making the reuse and implementation of those potentials in high-throughput searches significantly easier\cite{ko_materials_2025,cohen_torchsim_2025}. Both points contribute to a sustainable use of resources in the field.
Future work will need to devote itself to incorporating electronic properties of complex materials into such screening, such as by using machine-learned Hamiltonians\cite{li_critical_2025}. Again, data, code, and training infrastructure must be as open and reusable as possible to avoid multiple very similar developments in the field.

In the context of high-throughput searches, workflow tools and libraries such as atomate\cite{mathew_atomate_2017} and its successor atomate2\cite{ganose_atomate2_2025}, aflow\cite{curtarolo_aflow:_2012} have been developed. This allows easy reuse and extension of computational recipes in new high-throughput searches, making code development more sustainable in the field. While initially designed for high-throughput searches, they have now been extended to smaller-scale use cases, such as improving the reproducibility of calculations in the field. Additional workflow tools such as Aiida\cite{pizzi_aiida:_2016,huber_aiida_2020} and Pyiron\cite{janssen_pyiron_2019}, with initially different focuses (\eg provenance of simulations or an emphasis on more interactive workflow building), also exist. Recent advances in the field have aimed to bring these tools closer together, enabling easier reuse of the workflows\cite{janssen_python_2025}. Invaluable in this context have also been the pymatgen and ase software infrastructure, allowing simplified reuse of standard analysis and computations in the field\cite{ong_python_2013,larsen_atomic_2017}. The workflow tools are also nowadays used for building databases specifically for MLIPs, or for enabling iterative or active learning strategies for MLIPs\cite{menon_electrons_2024,liu_automated_2025,kuner_mp-aloe_2025,gharakhanyan_open_2025}. The enabled code reuse for different purposes contributes to a sustainable software infrastructure in the field.

With the help of classical high-throughput funnels, new materials for applications have been discovered (see summaries in Ref. [\citenum{horton_accelerated_2025}]). However, key challenges remain.

\textbf{Open challenges.} Some applications require properties that are extremely computationally demanding and complex to compute such as thermal conductivity\cite{park_advances_2025}, defect properties\cite{squires_guidelines_2025}. Significant advances have been made in automating these calculations using \textit{ab initio} methods and/or in combination with ML approaches (\eg MLIPs or supervised ML approaches optimized for smaller datasets\cite{de_breuck_materials_2021}). Despite these advances, the generation of reliable training data remains costly, making the open dissemination of databases, methods, and trained models essential to avoid redundant effort.

In many successful searches, already synthesized materials were used for screening. In this way, historical synthesis procedures can be reused, ensuring that the materials can be synthesized.  However, alternative structural candidates can also be generated through random structure searches\cite{pickard_ab_2011}, chemical-heuristic-inspired (substitution) algorithms\cite{hautier_data_2011}, global optimization algorithms\cite{wang_crystal_2010}, or, more recently, by generative approaches\cite{de_breuck_generative_2025,metni_generative_2025}. Evaluating such proposals introduces an additional challenge: synthesizability and viable synthesis pathways must be assessed alongside functional properties \cite{park_closing_2025}. To address this, researchers increasingly combine simulations with experimental data using approaches such as positive–unlabeled learning\cite{jang_structure-based_2020}, co-training strategies\cite{amariamir_syncotrain_2025}, literature mining\cite{kononova_text-mined_2019}, and sometimes in a combined manner\cite{chung_solid-state_2025}.
Simulations alone remain insufficient for this task. Additionally, the amorphous limit has been utilized to screen for both stable and metastable compounds\cite{aykol_thermodynamic_2018}.

Disorder is another significant challenge in computational searches for materials, as it is typically not accounted for in simulation-based databases\cite{leeman_challenges_2024}. This has been a major challenge for understanding and evaluating such searches. Disorder must be considered in both the generation of structural candidates and the evaluation of material properties. Recently, approaches have been introduced to incorporate disorder into generative models or to identify where it may need to be considered\cite{petersen_dis-gen_2025, jakob_learning_2025}. Also, experimental datasets to predict the formation of disordered phases have been collected as well\cite{lee_text-mined_2025}. Again, experimental datasets will need to be combined with simulations to make the consideration of disorder possible. Accessibility of experimental datasets will be crucial to advancing this field. Additionally, further microstructural effects (\eg grain boundaries) are crucial for the practical application of these materials. So far, they are rarely considered in high-throughput materials screening procedures based on atomistic simulations because they are challenging to simulate and characterize; yet, they play a central role in many material properties (\eg thermal conductivity\cite{hanus_thermal_2021}, elastic properties\cite{VOLKMER2026121844}). Promising for accurately capturing microstructure effects is the coupling of phase-field models with atomistic simulations (\eg as shown in Refs. [\citenum{wallis_linking_2026,wallis_linking_2026-1}]). 
Another major challenge is the inclusion of amorphous materials in high-throughput discovery\cite{liu_amorphous_2025}. Here, the development of efficient training and finetuning of MLIPs\cite{liu_automated_2025} or generative approaches for amorphous materials\cite{yang_generative_2025} will be critical, to make such simulations more sustainable. 

\begin{figure}[t]
    \centering
    \includegraphics[width=0.8\linewidth]{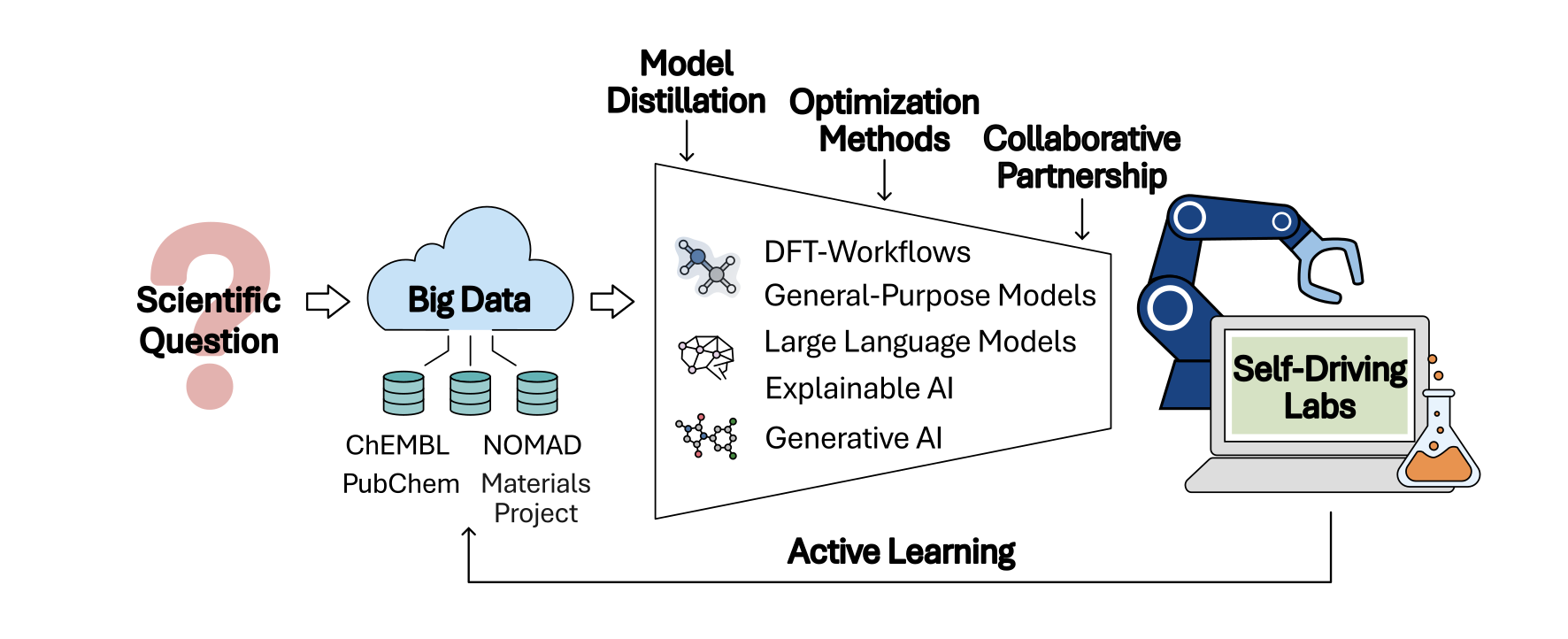}
    \caption{Scheme illustrating the desired workflow for sustainable exploration of chemical space in the discovery of technological materials and therapeutics. The process begins with the definition of a scientific question, followed by the navigation of existing big data repositories and the reuse of computational workflows and general-purpose models through collaborative partnerships. When implemented correctly, this approach can culminate in the development of self-driving laboratories capable of addressing the original question at larger scales.}
    \label{fig:last}
\end{figure}

\subsubsection{Toward autonomous discovery in self-driving labs} 

After seminal concept papers \cite{Tabor:2018aa,Aspuru2019} and initial successful demonstrations in 2019 \cite{doi:10.1126/sciadv.aaz8867, doi.org/10.1002/adma.201907801}, self-driving labs (SDLs) \cite{doi:10.1021/acs.chemrev.4c00055} for accelerated optimization of molecules, materials, processes, and devices are currently being developed in many academic and industrial labs worldwide. The ultimate goal of self-driving labs is maximization of efficiency to reach given objectives, \ie to find optimal molecular structures \cite{doi:10.1126/science.ads0901, doi:10.1126/science.adk9227}, synthesis conditions and procedures \cite{Shields2021}, materials compositions \cite{D3TA06651G}, or device architectures \cite{doi:10.1021/acs.accounts.4c00095} in the most efficient way. Efficiency can be defined in multiple ways, most of which relate to resource- and cost-efficiency, thereby increasing the sustainability of materials science and chemistry.

Platform development and (partial) automation typically take multiple years before platforms become productive \cite{Baird2022}. Step-by-step automation of critical bottleneck steps helps alleviate this challenge and leads to faster productivity and efficiency gains \cite{D3TA06651G, BAIRD2023102329}. Miniaturization \cite{doi.org/10.1002/smtd.202300553} and high-throughput optimized characterization and measurement methods \cite{doi:10.1021/acsnano.4c18760, doi.org/10.1002/elsa.202100122} lead to a further reduction of material consumption, but often go hand-in-hand with increased parallelization, which, at least to some extent, counterbalances the efficiency gains due to the creation of redundant data and a reduction of data efficiency.

Data efficiency is at the center of attention when it comes to decision-making algorithms used in SDLs. The most widely used algorithm is Bayesian optimization, while for some experimental setups, also genetic algorithms or search algorithms can play a role. Reinforcement learning has also been discussed, despite the fact that reinforcement learning is not an optimization algorithm but rather learns an optimal policy to act in an (uncertain) environment, which is a subtle difference to finding a good local or global optimum in a given search space in an efficient way.
Bayesian optimization \cite{frazier2018} is an optimization algorithm which combines adaptive learning with optimization. Bayesian optimization iteratively learns and updates a surrogate model of the input-output relations in the experiment, typically a Gaussian process regression model, and takes decisions based on optimizing a given acquisition function which balances between selecting promising experiments (exploitation) close to the presumed optimum of the surrogate model and experiments with unknown outcome (exploration) close to the maximum of uncertainty of the surrogate model. 
By doing so, Bayesian optimization aims to maximize data efficiency and, thus, indirectly also materials- and cost efficiency in SDLs. Due to the iterative nature of Bayesian optimization, parallelization reduces the data efficiency and thus sustainability, but potentially increases real-world-time-to-success, because an optimum might be found after fewer iterations and thus less overall real-world time (not to be confused with overall experimental time).
Despite its focus on data-efficiency and thus overall resource saving, Bayesian optimization has limitations which are not yet full overcome \cite{D3TA06651G}, including its lack of ability to learn from existing data and prior knowledge \cite{7822140, D3DD00177F, wang2025}, its rigidity towards fixed experimental workflows \cite{astudillo19a, NEURIPS2021_792c7b5a, torresi2025}, its efficiency regarding optimization problems with categorical variables \cite{GARRIDOMERCHAN202020, 10.1063/5.0048164}, as well as its general usability \cite{baird2025, doi:10.26434/chemrxiv-2025-f1wcr}.

\textbf{Open challenges.} From the perspective of data efficiency and sustainability, solving the cold-start challenge of Bayesian optimization and allowing the decision-making algorithm to incorporate prior data from related optimization campaigns, either with the same variables or even with different search space dimensions or objectives, would have a major impact on the cost and speed of optimization. However, current approaches can also mislead the surrogate model and lead to suboptimal convergence \cite{RODEMANN2024112186, li2026robust, D5DD00337G}. Depending on how relevant and meaningful the prior data is, the required number of experimental iterations would be significantly decreased, all the way to few- or zero-shot optimizations. The role of new types of surrogate models and acquisition functions \cite{NEURIPS2023_2561721d}, as well as LLMs \cite{wang2025} in achieving this is yet to be determined. Further improvements will be made possible by moving from fixed SDL workflows, which model the entire experimental workflow as one process with input variables and observable objectives, to decision-making algorithms, which model the experimental workflow as a function network, where intermediate observations ("proxy measurements") can be used to reduce the overall number of experimental steps needed \cite{NEURIPS2021_792c7b5a, torresi2025}. Transfer between search spaces as well as the data-efficient exploration of more complex search spaces requires efficient handling of categorical variables, which typically requires the definition of problem-specific features representations of each instance, which is straight forward for a finite and small list of instances (\eg a small number of possible solvents \cite{doi.org/10.1002/anie.202200242}) but increasingly hard for large or even infinite categorical search spaces (\eg exploring molecular space \cite{doi:10.1126/science.ads0901}). Approaches to learn such representations on-the-fly or using general-purpose models could be an important milestone toward even further increasing the data-efficiency and thus sustainability of SDL-based materials discovery.

\section{Concluding remarks}

The emergence of general-purpose ML models is transforming computational materials science and chemistry by enabling transfer across broad chemical spaces and substantially reducing the need for system-specific training and data generation. From a sustainability perspective, the considerable energy required to produce large datasets and general-purpose models can only be justified through open dissemination and widespread reuse. In this sense, multi-fidelity strategies that combine different levels of theory, together with model distillation and efficient inference architectures, will allow these models to be deployed in computationally tractable forms with minimal loss of accuracy. Moreover, in hierarchical workflows, fast ML surrogates can be applied broadly while expensive QM methods are invoked selectively for electronically complex cases, thereby optimizing the allocation of computational resources.

A key frontier towards a more sustainable exploration of chemical spaces is the integration of physics-based knowledge and rigorous sampling into ML-driven modeling. Embedding conservation laws, symmetries, dimensional consistency, and other physical constraints directly into model construction improves generalization, interpretability, and data efficiency. At the same time, bridging the gap between idealized training data--often derived from zero-temperature density functional theory--and real-world conditions requires accounting for temperature, disorder, finite-size effects, and synthesis constraints. Generative AI frameworks should therefore incorporate multi-objective criteria such as elemental abundance, toxicity, recyclability, and synthesizability to avoid impractical proposals. Accordingly, closing the loop between computation and experiment through automated laboratories capable of rapid validation represents a crucial step toward sustainable material and molecular discovery.

Ultimately, sustainable progress will depend on openness, modular workflows, and efficient use of both data and computation (see Fig. \ref{fig:last}). Shared datasets, models, and infrastructure prevent redundant effort, while smaller domain-specific models and agentic AI workflows can deliver high utility at relatively low energy cost. Active learning, automation, and adaptive optimization strategies further reduce the number of required simulations and experiments, particularly when prior knowledge and intermediate observations are exploited. Extending these approaches to complex systems, including microstructures, metamaterials, intrinsically disordered proteins, and biomolecular condensates, remains a major challenge. As ML-driven methods become mainstream, their success should be judged not only by predictive accuracy but also by their capacity to reduce computational cost, experimental waste, and environmental impact while enabling sustainable discovery of materials and therapeutics. In this spirit, the SusML workshop--and this Perspective--underscore the collective efforts of the computational science community to develop resource-efficient strategies for exploring chemical space using machine learning.

\section*{Author contributions} 

L.M.S., M.T., A.T., M.R., and G.C secured the funding and organized the SusML workshop. L.M.S., A.T. and M.R. noted topics of interest for the perspective during the workshop. 
L.M.S. and M.R. authored the introduction.
A.B., T.F., Y.L., and J.T.M. authored the section ``Equivariant machine learning force fields''.
L.M.S. and A.H.C. authored the section ``Quantum-informed representations for drug discovery''.
L.G. authored the section ``Parsimonious models for materials informatics''.
R.F. authored the section ``Data-driven design of non-van der Waals two-dimensional materials''.
W.D. authored the section ``High-accuracy electronic structure \& correlated materials''.
D.B. and H.K. authored the section ``Generative AI for molecules''.
H.T. authored the section ``Generative AI for inorganic materials''.
A.E. authored the section ``Language models as predictive tools''.
J.M.C. authored the section ``Language models as extractive tools''.
V.J. authored the section ``Uncertainty-aware exploration of energy landscapes''.
M.T. authored the section ``Sustainable materials sampling with costly simulations''.
V.L.D. authored the section ``Data efficiency in machine-learned interatomic potentials''.
J.G. authored the section ``High-throughput discovery in inorganic materials''.
P.F. authored the section ``Toward autonomous discovery in self-driving labs''.
L.M.S. coordinated the sections, the figures, and revised the manuscript organization.
All authors revised the final manuscript.

\section*{Acknowledgements} 

L.M.S., M.T., A.T., M.R., and G.C would like to thank the International Office at the TUD Dresden University of Technology, the Psi-k community, RSC Digital Discovery, and the Max Planck Institute for the Physics of Complex Systems for the financial and institutional support to organize the SusML workshop. 
L.M.S. and G.C. gratefully acknowledge the funding by the German Research Foundation (DFG) under the Cluster of Excellence CeTI: Centre for Tactile Internet with Human-in-the-Loop (EXC 2050/2, Project ID 390696704), Cluster of Excellence REC²: Responsible Electronics in the Climate Change Era (EXC 3035), and Cluster of Excellence CARE: Climate-Neutral And Resource-Efficient Construction (EXC 3115, Project ID 533767731). 
H.K. and D.B. acknowledge the RCN programs for ground-breaking research (CatLEGOS grant 325003) and National Centers of Excellence (Hylleraas grant 262695), and the Norwegian Supercomputing Program (grants NN4654K and NS4654K).
W. D. acknowledges funding from the German Federal Ministry of Research, Technology and Space (BMFTR) via the research program Quantensysteme and funding measure Quantum Futur 3 for project No. 13N17229 as well as the Helmholtz Association via Initiative and Networking Fund for projects No. VH-NG-21-08, KA-QUS-02 (qFLOW) and KA-QUS-03 (QT-Batt).
V.L.D. acknowledges support from UK Research and Innovation [grant number EP/X016188/1].

\bibliography{ref_susml}

\end{document}